\documentclass[runningheads]{llncs}

\usepackage{eccv}

\usepackage{eccvabbrv}

\usepackage{graphicx}
\usepackage{booktabs}
\usepackage{wrapfig,lipsum,booktabs}
\usepackage{multirow}
\usepackage[table]{xcolor}
\usepackage{comment}
\usepackage{multicol}
\usepackage{caption}
\usepackage{placeins}
\usepackage{hyperref}
\usepackage[accsupp]{axessibility}  

\usepackage{hyperref}

\usepackage{orcidlink}

\usepackage{amssymb} 
\newcommand{\cmark}{\checkmark}
\newcommand{\xmark}{\texttimes}

\begin{document}

\title{Learning to Deny: Action Denial in Multimodal Large Language Models} 


\author{Raiyaan Abdullah\orcidlink{0000-0002-7950-816X} \and
Shehreen Azad\orcidlink{0009-0003-4675-3817} \and
Yogesh Singh Rawat\orcidlink{0000-0003-4052-6798}}

\authorrunning{R.~Abdullah et al.}

\institute{Institute of Artificial Intelligence, University of Central Florida, USA\\
\email{raiyaanabdullah@gmail.com}, \email{shehreen.azad@ucf.edu}, \email{yogesh@crcv.ucf.edu}\\
Project page: \url{https://raiyaan-abdullah.github.io/Learn-to-Deny-webpage}
}

\maketitle

\begin{abstract}
Multimodal large language models (MLLMs) have rapidly advanced video understanding, achieving strong zero-shot and few-shot recognition across standard benchmarks. Yet their ability to \textbf{deny} an action by recognizing when an activity is \textbf{not} happening despite strong contextual cues remains largely unexplored. We introduce \textbf{UCF101-AD}, a large-scale benchmark consisting of paired \textit{Action-Presence} and \textit{Action-Denial} clips, designed to evaluate this capacity for denial. Each negative video in UCF101-AD preserves the same contextual and motion cues (persons, objects, locations) as its positive counterpart, but the defining action itself is explicitly absent. Evaluating 20 state-of-the-art MLLMs reveals a consistent failure: models that exceed 85\% accuracy on the positive action classes collapse below 50\% on its action-denial counterpart, indicating a strong inclination to affirm plausible actions rather than verify that they truly occur. This exposes a critical blind spot in modern video understanding: the inability to reason causally about whether a motion actually happens. To probe this issue, we explore a causal graph formulation, \textbf{CausalAct}, which expresses scene structure through natural-language prompts linking context, interaction, and motion. Incorporating such causal cues substantially reduces false positives, demonstrating that denial is a learnable reasoning skill. UCF101-AD provides a new lens for diagnosing and improving causal reasoning in multimodal models. Dataset and relevant code: \url{https://github.com/raiyaan-abdullah/Learn-to-Deny}.
\end{abstract}    
\section{Introduction}
\label{sec:intro}

\begin{figure}[t!]
\centering
{
  \includegraphics[width=1\textwidth]{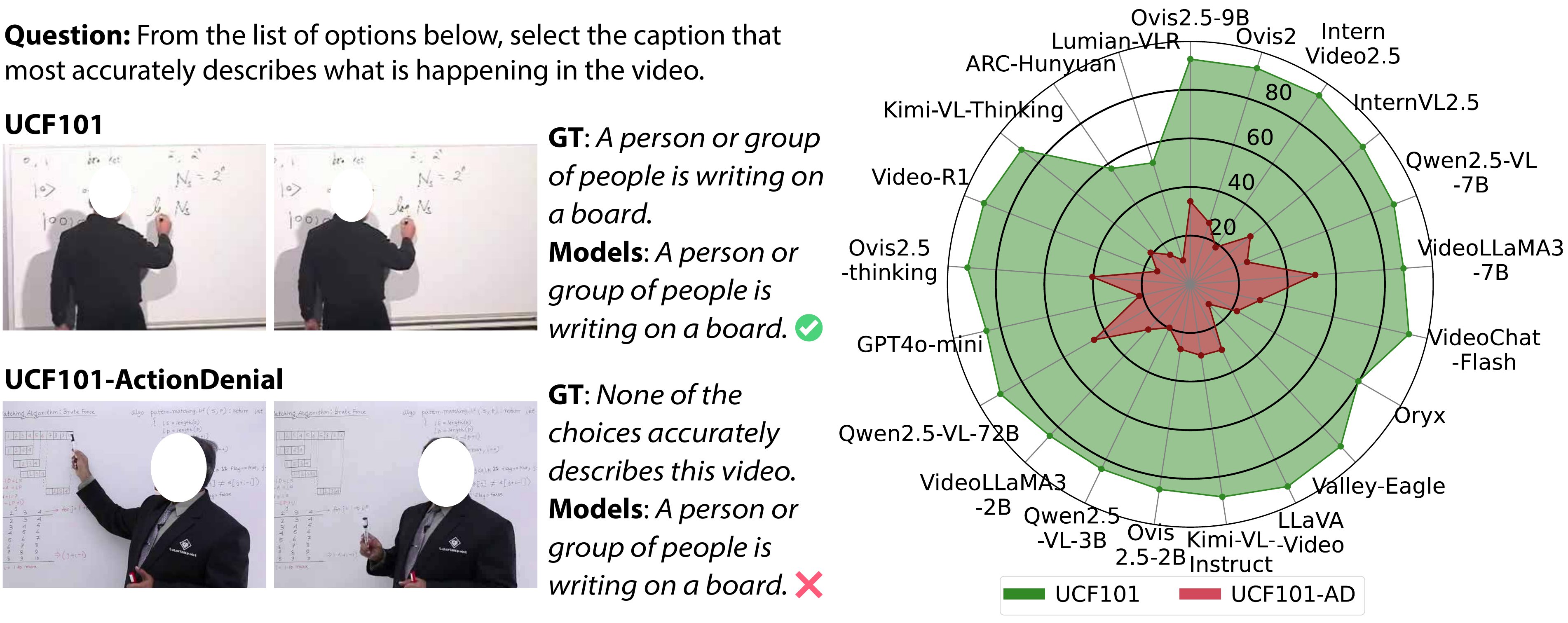}
}
\caption{\textbf{\textit{Model failures in denying an action:}} \textit{(left)} Models are capable of recognizing when the action occurs, but seeing a person holding a marker in front of a board but not writing, they still predict ``Writing On Board''. \textit{(right)} Radar plot of accuracy on the original UCF101 test set (green) vs. the UCF101-AD \textit{Action-Denial} videos (red). Most models exceed 85\% on UCF101 but fall below 50\% on videos where the action is absent, highlighting a large gap between action recognition and denial.
}
\label{fig:teaser}
\end{figure}

Recent Multimodal Large Language Models (MLLMs) \cite{actionclip, vificlip, xclip, froster, t2l, internvideo2_5} achieve strong results on standard action recognition benchmarks, but closer analysis shows that much of this performance can arise from spurious correlations rather than genuine temporal reasoning \cite{punching_bag_vs_person, what_makes_video_video, dynamic_spatio_tempo, masking_bg_obj, albar}. Instead of verifying the motion that defines an activity, models often rely on contextual cues such as background or object presence. This aligns with evidence that MLLMs display an agreeableness or affirmative bias, tending to endorse a premise rather than negate it \cite{sycophancyllms, sycophancyllms2, syceval}. In action recognition, this bias appears as predicting that an action is occurring whenever context is suggestive, even if the defining motion is missing, leading to false positives and potential failures in settings like surveillance, autonomous systems, and sports analytics. We formalize this challenge as \textbf{action denial}: the ability of a model to determine that an action is not occurring when its defining motion is absent, despite compelling contextual evidence (\cref{fig:teaser}, \textit{left}). Robust video understanding therefore requires models to confirm actions through their defining motion and to reliably deny them when that motion is not observed.

Despite the scale and diversity of benchmarks such as Kinetics \cite{kinetics}, UCF101 \cite{ucf101}, HMDB51 \cite{hmdb51}, and ActivityNet \cite{activitynet}, they primarily contain positive examples where the labeled action truly occurs and its defining motion is visible. They rarely include hard counterfactual negatives in which relevant objects, actors, or scenes are present but the action does not take place. This reinforces object-action correlations and allows models to succeed without explicitly verifying motion. Since models are neither trained nor evaluated on contextually similar yet motion-absent clips, current benchmarks offer limited insight into whether a system can separate genuine action evidence from misleading context and reliably deny an action when its defining motion is missing.

To address this gap and directly evaluate action denial, we introduce \textbf{UCF101-ActionDenial (UCF101-AD)}, a dataset of 11,283 videos built on UCF101 \cite{ucf101} action categories, with paired positive and negative classes for each action. Each negative clip preserves the typical context of its positive counterpart while ensuring the target action does \textbf{not} occur, and includes realistic motion that often co-occurs with the action. For instance, in ``Not Basketball Dunk,'' players dribble on a court with a ball present, but no dunk happens. This controlled design isolates defining motion as the key discriminative signal and provides a focused benchmark for evaluating action denial.

Our experiments reveal that the \textbf{capacity for denial}, a fundamental reasoning skill, is exceptionally challenging for multimodal action models and general-purpose MLLMs. We observe that several state-of-the-art architectures struggle to differentiate these negatives, most achieving performance below 50\%, often confusing with the positive action (\cref{fig:teaser}, \textit{right}). This failure highlights that models are learning spurious correlations rather than the true causal relationship between interaction, motion and target action. They lack a reliable mechanism to verify if the defining interaction is present.

We hypothesize that models can more reliably decide whether an action is occurring, including its absence, if they are guided to reason about how contextual evidence (persons, objects, location) supports the defining motion. Existing debiasing methods that weaken correlations with background or object cues \cite{dance_in_the_mall, mitigating_representation,scene_debiasing_open,albar} operate mainly in positive-action regimes and reward correct labels, not explicit denial when motion is missing. Robust denial instead calls for a structured representation that organizes scene components and checks whether the preconditions of an action are actually instantiated. Prior work on causal structure and spatio-temporal scene graphs shows that modeling dependencies among entities, relations, and events can improve robustness \cite{vcdn,mecd,actiongenome}, but is typically used for prediction or explanation. In contrast, we ask whether an explicit structure can help models decide when to deny a target action despite strong contextual cues. 

Motivated by this, we propose \textbf{CausalAct}, a causal framework that organizes scene components and their dependencies with a focus on verification. Building on graph-based prompting and structured scene descriptions that improve multimodal video reasoning \cite{graphprompts,sgvlm,causalgraphexperts}, we express the causal graph in natural language and ask the models to check whether the necessary conditions for the target action hold. We find that this improves action-denial performance for models with strong language backbones, indicating that the language module must be able to parse such graph-style prompts. We further finetune smaller MLLMs on auxiliary graph-related VQA tasks that teach graph concepts without exposing target action labels, avoiding label leakage, and observe similar gains, showing that causal graphs can aid denial once the model learns to interpret them. Our contributions are:

\begin{itemize}
  \item We introduce \textbf{UCF101-AD}, a benchmark for action denial reasoning where the scene context and motion cues remain intact but the target action is absent, comprising 11,283 video samples. 
  
  \item We evaluate \textbf{20 state-of-the-art MLLMs}, demonstrating that their strong action recognition capabilities fail to translate to robust action denial when misleading contextual cues are present. 

  \item We propose \textbf{CausalAct}, a graph-based finetuning framework that represents scene structure as natural language graphs and trains models through graph-centric VQA tasks to improve reasoning about action absence.
\end{itemize}

\section{Related Work}
\label{sec:relwork}
\noindent\textbf{Video Action Recognition Benchmarks:} Action recognition datasets span trimmed clips \cite{ucf101, hmdb51, kinetics, ssv2, momentsintime}, untrimmed videos \cite{activitynet, thumos14, hacs}, and domain-specific settings such as sports, instructional content, and driving \cite{sports1m, diving48, finegym, soccernet, dada, howto100m}, as well as egocentric views \cite{epic-kitchens, ego4d}, multi-label action understanding \cite{ava, ava-kinetics, charades, multithumos, isafetybench}, and skeleton/depth/RGB-based benchmarks \cite{ntu-rgbd, pku-mmd, sbu-kinect}. Most of these focus on recognizing positive actions; negatives are usually implicit, e.g., background or unlabeled frames \cite{hacs, thumos14}, or are used to probe related failure modes such as similar same-scene actions, missing context, or failed executions \cite{cdad, mimetics, oops}. \cite{grouplet} addresses related co-occurrence ambiguities in still images. However, there remains a gap to test denial of a target action in videos when familiar context, objects, or motion cues are present but the defining action is absent.

\medskip
\noindent\textbf{Causal Reasoning:} Work on causal reasoning for multimodal models includes synthetic benchmarks with ground-truth graphs for probing interventions and temporal dependencies \cite{cleverer, causalcity, vcdn}, and large-scale annotated datasets for real-world event-level reasoning \cite{mecd, actiongenome}. Graph-based approaches use scene graphs or structured prompts to organize visual content for multimodal reasoning \cite{graphprompts, sgvlm, hyperglm, causalgraphexperts, hdepickg, mvuframework}, while other efforts ask whether LLMs can internalize abstract causal structures and use them for robust inference \cite{cladder, c2p, clear, causalprompting, axiomatictraining}. Building on these, our work decomposes videos into structured scene components and models their relations, specifically evaluating whether MLLMs can decide if a target action truly occurs and deny it when the proposal conflicts with the causal evidence.

\medskip
\noindent\textbf{Multimodal Large Language Models (MLLMs):} Video-language research has progressed from retrieval and contrastive pretraining \cite{clip4clip, videoclip, openvclip, centerclip, xpool} to temporal action models \cite{xclip, vificlip, actionclip, froster, t2l}, and more recently to LLM-centric vision-language systems \cite{internvl2_5, qwen-vl2_5, internvideo2_5, ovis2_5} and instruction-tuned video MLLMs \cite{videollama3, llavavideo, videochat, valley-eagle}. Emerging ``thinking'' models add explicit reasoning \cite{deepseek, arc-hunyuan, lumian-vlr, video-r1, kimi-vl}, but many are tuned in ways that induce agreeableness or sycophancy \cite{mitigatingagreeablenessbias, sycophancyllms, openaisycophancy, syceval, chaoskeywords, sycophancyllms2}. While these works study bias and reasoning in isolation, evaluation of causal reasoning for action denial in videos remains absent, leading to a dual failure: models are biased by misleading context and prone to affirming false positives. Our work provides a first systematic assessment of this denial capacity and a method to guide MLLMs toward robust, non-sycophantic action understanding.

\section{Benchmarking Action Denial}
\begin{figure}[]
    \centering
    \includegraphics[width=\linewidth]{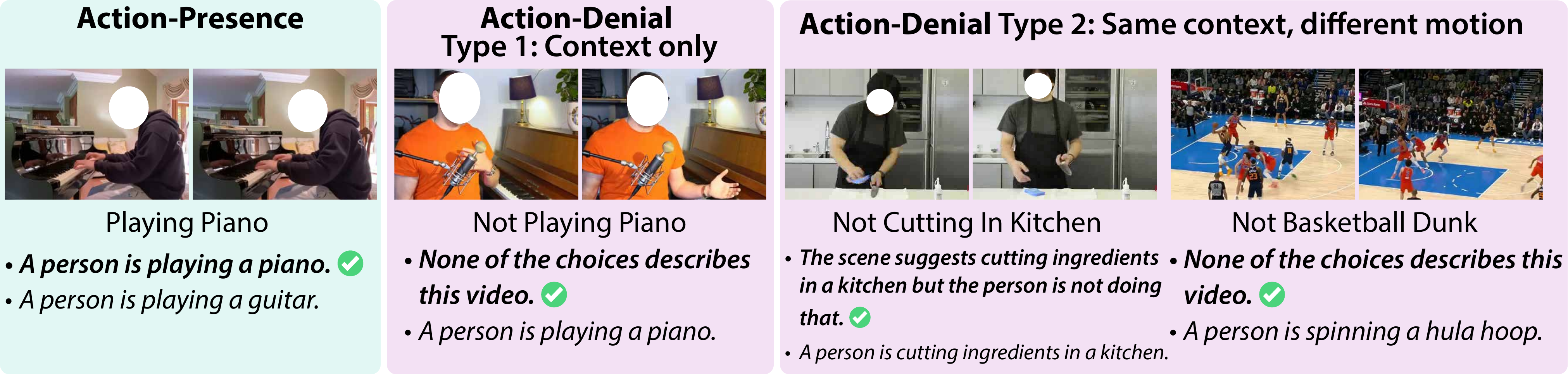}
    \caption{\textbf{Overview of UCF101-AD.} The dataset contains \textit{Action-Presence} clips and hard \textit{Action-Denial} negatives. Negatives come in two types: \textit{\textbf{Type 1 - Context only}}, where the usual scene and objects are present but the defining motion is absent, and \textit{\textbf{Type 2 - Same context, different motion}}, where a different action occurs in the same setting. For the task where the model must select the correct caption for the video, we show examples with different MCQ setups; answer choices are shortened and subsampled for display.
    }
    \label{fig:ucf101_neg_example}
\end{figure}

To evaluate a model’s ability to deny actions under strong contextual cues, we propose \textbf{UCF101-ActionDenial (UCF101-AD)}, a benchmark derived from UCF101 action classes~\cite{ucf101}. The dataset contains both \textit{Action-Denial} and \textit{Action-Presence} clips. \textit{Action-Denial} videos preserve scenes, objects, and actors associated with a target class, but the defining motion of that action does not occur. These include cases where no defining motion is present or where a different motion occurs despite similar context. This design creates visually plausible clips where context suggests an action while the action itself does not take place, directly testing a model’s ability to recognize action absence. \textit{Action-Presence} videos contain the defining motion of the target action and serve as a reference for standard action recognition. An overview of the benchmark is shown in \cref{fig:ucf101_neg_example} with \cref{tab:benchmark_comparison_new} showing comparison with prior benchmarks. While prior works primarily treat negatives as localization distractors, failed attempts, or contextually ambiguous clips, UCF101-AD is explicitly designed around action denial: negatives are constructed to remove the defining motion while preserving plausibly misleading context, and are paired with positive counterparts to support structured reasoning about absence.

\begin{table}[t]
    \scriptsize
    \centering
    \captionof{table}{\textbf{Comparison of UCF101-AD with prior benchmarks.} No target motion indicates classes where the defining motion is absent. Pos-Neg Pair indicates whether each action has an explicit positive and negative counterpart. \cmark$^*$ denotes applicability to a subset of denial clips.
    }
     \label{tab:benchmark_comparison_new}
    \setlength{\tabcolsep}{3pt}
    \resizebox{\linewidth}{!}{
    \begin{tabular}{l|cccc|cccc}
     \toprule
     \multirow{3}{*}{\textbf{Benchmark}} & \multicolumn{4}{c|}{\textbf{Negative Composition}} & \multicolumn{4}{c}{\textbf{Negative Structure}}\\
     &  \multirow{2}{*}{\textbf{Scope}} & \multirow{2}{*}{\textbf{Coverage}} & \multirow{2}{*}{\textbf{\% Clips}} & \multirow{2}{*}{\textbf{Semantics}} & \textbf{No Target} & \textbf{Pos-Neg} & \textbf{Explicit} &\textbf{Reasoning} \\
     &  & & & & \textbf{Motion} & \textbf{Pair}& \textbf{Annotation} & \textbf{Challenge} \\
     \midrule  
     HACS \cite{hacs}   & Frame & Partial & - & Distractor & \xmark & \xmark & \xmark & Action localization \\
     THUMOS14 \cite{thumos14}  & Frame & Partial & - & Distractor & \xmark & \xmark & \xmark &Action localization\\
     CDAD \cite{cdad} & Frame & Partial & - & Distractor &\xmark & \xmark & \xmark & Action localization \\
     OOPS \cite{oops} & Frame & Partial & 100 & Attempt &\xmark & \xmark & \cmark  & Failure detection \\
     Mimetics \cite{mimetics}  & Clip & Subset & - & Contextual & \xmark & \xmark & \xmark & Spatial ambiguity \\
     SSv2 \cite{ssv2} & Clip & Subset &10& Attempt & \xmark & \cmark$^*$ & \cmark  & Outcome detection\\
     \midrule
     {\textbf{UCF101-AD}}  & {\textbf{Clip}} &{\textbf{Subset}} & {\textbf{94}} & {\textbf{No action}} &{\textbf{\cmark}} & {\textbf{\cmark}} & {\textbf{\cmark}}& \textbf{Action Denial} \\
     \bottomrule
    \end{tabular}
    }   
\end{table}

\subsection{UCF101-AD Construction}
\noindent \textbf{Video Curation:}
We construct the negative \textit{Action-Denial} videos by collecting clips corresponding to each UCF101 action class where the defining action does not occur while preserving similar contextual cues such as scene, objects, or actor appearance. These clips resemble the typical setting of the target action but lack the characteristic motion that defines it. Here, the \textit{defining motion} refers to temporally extended activities whose recognition requires modeling motion dynamics over time. Accordingly, we focus on actions characterized by such temporal patterns rather than atomic states such as sitting, standing, or listening. Moreover, the presence of an action requires the full execution of its characteristic motion pattern; partial or incomplete movements are not considered sufficient evidence. For instance, slightly moving a golf club does not constitute a golf swing unless the complete swing motion is performed. Based on how the absence of the target motion manifests, we divide the \textit{Action-Denial} videos into \textbf{two types}:
\begin{enumerate}
    \item \textbf{Context only:} Videos preserve the salient objects and environment associated with the target action, but no defining interaction takes place. For example, ``Not Playing Piano'' shows a person near a piano without actually playing it.
    \item \textbf{Same context, different motion:} Videos retain the same setting and object interactions as the target action, but the actor performs other plausible actions while never executing the defining motion. For example, ``Not Basketball Dunk'' shows players dribbling, passing, or shooting on a basketball court, yet no dunk occurs.
\end{enumerate}
Together, these two negative categories disentangle contextual cues from motion evidence, requiring models to verify the defining motion rather than rely on correlated context. In contrast, the positive \textit{Action-Presence} videos contain clips where the defining motion occurs, serving as a reference to verify that models correctly recognize actions when they are actually performed.
All videos are sourced from YouTube and temporally segmented into short clips. Following  UCF101~\cite{ucf101}, clips originating from the same source video are assigned entirely to either the training or test split to avoid information leakage.

\noindent \textbf{QA Generation.} For each video, we generate a question asking which description best matches the content of the video, along with several candidate options in a multiple-choice format. We employ this fixed multiple-choice-question based VQA setup to enable objective evaluation by avoiding the variability of free-form generation \cite{seedbench, mvbench}. For \textit{Action-Denial} videos, the answer choices include a diverse set of candidate actions, including the main distractor corresponding to the original UCF101 target action, additional actions sampled from other categories, and a ``None” option. This setup introduces strong visual and linguistic cues toward the target action, allowing us to evaluate whether models can correctly deny the action despite these cues. For \textit{Action-Presence} videos, the correct answer is the target action, and the remaining options are closely related actions plus the same ``None'' option, forcing the model to distinguish the specific motion rather than rely on coarse context.

\noindent\textbf{Quality control.} 
We employ a fully human verification pipeline to ensure dataset reliability. Annotators first collect candidate videos through keyword searches targeting relevant context without the target action for the \textit{Action-Denial} set. Each clip is manually verified to ensure sufficient contextual cues and the complete absence of the corresponding action, including partial executions. Low-quality or irrelevant clips are discarded. For the \textit{Action-Presence} set, only clips containing a clear instance of the action are retained. All clips are cross-validated by annotators, removing about 20\% of candidates. No MLLMs were used in this process to avoid model-induced bias. After this process, UCF101-AD contains 11,283 clips in total: the \textit{Action-Denial} split includes 7,059 negative clips for training and 3,549 for testing, while the \textit{Action-Presence} split consists of a test-only set of 675 clips with no training clips. In QA generation for every distractor choice, the original UCF101 label is rewritten as a clear descriptive sentence to reduce ambiguity; for example, ``HulaHoop'' becomes ``A person or group of people is spinning a hula hoop.''

\subsection{Benchmarked Models}
We evaluate \textbf{20} representative state-of-the-art open-source MLLMs on the test splits of \textbf{UCF101-AD} to assess their capacity for action denial. All of our chosen models are trained on very large (public and/or proprietary) datasets.

\noindent \textbf{General-Purpose Models.} The evaluated models span major families including Ovis \cite{ovis, ovis2_5}, Intern \cite{internvideo2_5}, Qwen \cite{qwen-vl2_5}, and VideoLLaMA \cite{videollama3}, alongside other key architectures \cite{valley-eagle, oryx, videochat, llavavideo}; differing in training scale and alignment strategies. We also include the closed-source model GPT-4o mini \cite{gpt4omini}.

\noindent \textbf{Reasoning Models.}
To examine whether explicit multi-step inference improves action denial, we additionally evaluate recent reasoning-oriented MLLMs, including Ovis 2.5 \cite{ovis2_5}, Video R1 \cite{video-r1}, ARC-Hunyuan \cite{arc-hunyuan}, Kimi-VL \cite{kimi-vl} and Lumian-VLR \cite{lumian-vlr}. These models generate intermediate reasoning steps before producing a final prediction, enabling more systematic verification.

\subsection{Implementation Detail, Evaluation Protocol and Metric} 
All models are evaluated using their official implementation and released weights. In the multiple-choice (MCQ) setting, the order of the answer options is randomized, and the ground-truth answer is assigned to a random position among them. Each question contains 11 options in total. which ensures that the prompt is reliably processed by all MLLMs while limiting the effectiveness of random guessing to $<10\%$.

We evaluate all models in a visual question answering (VQA) setup and report several accuracy metrics. For negative clips, we measure \textit{Type 1} and \textit{Type 2} accuracy and their size-weighted average, \textit{Overall Action-Denial (Overall-AD)}. For positive clips, we report \textit{Action-Presence} accuracy. We also provide an \textit{Overall} accuracy as the size-weighted average of \textit{Overall-AD} and \textit{Action-Presence}. To assess a model’s balanced ability to both recognize and deny actions, independent of class sizes, we report the \textit{Harmonic Mean (HM)} of \textit{Overall-AD} and \textit{Action-Presence}.

\begin{table}[]
\scriptsize
\centering
\caption{\textbf{Benchmarking models on UCF101-ActionDenial.} Reported metric is accuracy (\%) for \textit{Type 1} negatives, \textit{Type 2} negatives, \textit{Overall-AD} (Action-Denial) accuracy, \textit{Action-Presence} accuracy, and raw \textit{Overall} dataset accuracy. The last column reports the \textit{Harmonic Mean (HM)} of the \textit{Overall-AD} and \textit{Action-Presence} accuracies. \textbf{Best} and \underline{second-best} performance are highlighted.
}
\label{tab:ucf101neg_results}
\resizebox{\linewidth}{!}{
\begin{tabular}{l | c c c| c| c c}
\toprule
 & \multicolumn{3}{c|}{\textbf{Action-Denial}} & \textbf{Action-} & & \\
\textbf{Model} & \textbf{Type 1} \textuparrow & \textbf{Type 2} \textuparrow & \textbf{Overall-AD} \textuparrow & \textbf{Presence} \textuparrow & \textbf{Overall} \textuparrow & \textbf{HM}  \textuparrow \\ 
\midrule
Ovis2.5-9B \cite{ovis2_5}                  & 34.5 & 33.8 & 34.1 & \underline{97.5} & 43.7 & 50.5 \\
Ovis2-8B \cite{ovis}                       & 25.6 & 27.1 & 26.4 & 95.6 & 36.9 & 41.4 \\
InternVideo2.5\_Chat-8B \cite{internvideo2_5} & 19.1 & 17.7 & 18.4 & 96.6 & 30.3 & 30.9 \\
InternVL2.5-8B \cite{internvl2_5}          & 30.5 & 32.5 & 31.6 & 93.5 & 41.0 & 47.2 \\
Qwen2.5-VL-7B-Instruct \cite{qwen-vl2_5}   & 22.0 & 27.9 & 25.1 & 95.4 & 35.8 & 39.7 \\
VideoLLaMA3-7B \cite{videollama3}          & \textbf{49.4} & \textbf{53.4} & \textbf{51.5} & 96.0 & \textbf{58.3} & \textbf{67.0} \\
VideoChat-Flash-Qwen2-7B \cite{videochat}  & 27.4 & 31.1 & 29.4 & 94.4 & 39.3 & 44.8 \\
Oryx-7B \cite{oryx}                        & 20.6 & 23.5 & 22.1 & 85.6 & 31.7 & 35.1 \\
Valley-Eagle-7B \cite{valley-eagle}        & 8.8  & 13.1 & 11.1 & 96.4 & 24.0 & 19.9 \\
LLaVA-Video-7B-Qwen2 \cite{llavavideo}     & 28.9 & 30.7 & 29.9 & 96.4 & 40.0 & 45.6 \\
Kimi-VL-A3B-Instruct \cite{kimi-vl}        & 27.4 & 31.4 & 29.6 & 94.7 & 39.5 & 45.1 \\
Ovis2.5-2B \cite{ovis2_5}                  & 23.1 & 30.4 & 27.0 & 91.1 & 36.7 & 41.7 \\
Qwen2.5-VL-3B-Instruct \cite{qwen-vl2_5}   & 17.2 & 22.1 & 19.8 & 92.3 & 30.8 & 32.6 \\
VideoLLaMA3-2B \cite{videollama3}          & 20.2 & 30.0 & 25.4 & 93.6 & 35.8 & 40.0 \\
Qwen2.5-VL-72B-Instruct \cite{qwen-vl2_5}  & \underline{42.6} & \underline{47.8} & \underline{45.7} & \textbf{97.6} & \underline{53.6} & \underline{62.3} \\
GPT4o-mini \cite{gpt4omini}                                & 20.7 & 22.3 & 21.5 & 90.1 & 31.9 & 34.7 \\
\midrule
\multicolumn{7}{c}{\textit{Reasoning Models}} \\
\midrule
Ovis2.5-9B (thinking) \cite{ovis2_5}       & 36.7 & 43.8 & 40.4 & 96.7 & 48.9 & 57.0 \\
Video-R1-7B \cite{video-r1}                & 12.7 & 16.3 & 14.6 & 96.0 & 27.0 & 25.3 \\
Kimi-VL-A3B-Thinking \cite{kimi-vl}        & 14.9 & 26.1 & 20.9 & 96.7 & 32.4 & 34.4 \\
ARC-Hunyuan-Video-7B \cite{arc-hunyuan}    & 12.9 & 16.3 & 14.7 & 58.5 & 21.3 & 23.5 \\
Lumian-VLR-7B-Thinking \cite{lumian-vlr}   & 9.8  & 10.7 & 10.3 & 55.3 & 17.1 & 17.4 \\
\bottomrule
\end{tabular}
}
\end{table}

\subsection{Results}
\cref{tab:ucf101neg_results} summarizes zero-shot performance of all models on UCF101-AD. Across models, \textit{Action-Presence} accuracy is generally high (often above 90\%), confirming that current MLLMs recognize actions reliably when the motion is present. In contrast, \textit{Action-Denial} accuracy is much lower: even the strongest model, VideoLLaMA3-7B, reaches only 51.5\% \textit{Overall-AD} on negatives, and most models fall in the 20-35\% range. Type~1 negatives, which preserve the canonical objects and environments but do not contain the motion, are harder for most models, suggesting that they heavily over-index on context and object presence rather than verifying the critical motion. Type~2 negatives, which keep the same setting but replace the target motion with other plausible activities, yield slightly higher scores for most models, yet performance remains far from robust, implying that models often fail to distinguish fine-grained motion patterns that disambiguate the target action from related activities. As a result, the \textit{Overall} and \textit{Harmonic Mean} between \textit{Overall-AD} and \textit{Action-Presence} accuracy remain modest (maximum 58.3 and 67.0), showing that strong action recognition alone does not translate into robust action denial under strong contextual cues.

\noindent \textbf{Human evaluation of action denial}. To assess whether the action-denial examples are reliably verifiable by humans, we conducted a human evaluation on UCF101-AD with 36 participants. Each participant viewed videos from five randomly sampled denial classes and judged whether the target action was absent. Participants achieved 86.6\% \textit{Overall-AD}, indicating that humans can reliably identify action absence even under misleading contextual cues.

\subsection{Analysis}

\begin{figure}[]
    \centering
    \includegraphics[width=\linewidth]{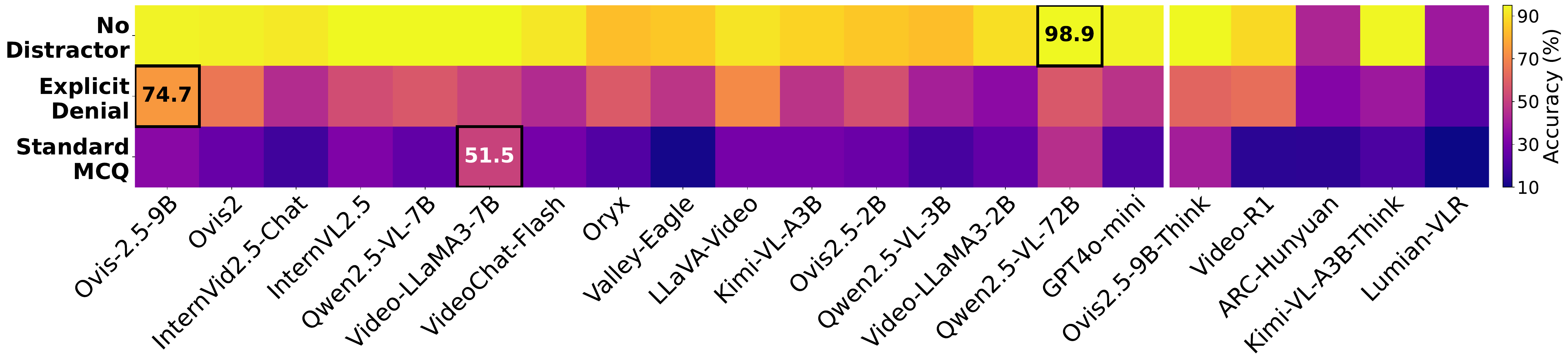}
    \caption{\textbf{Effect of progressive hinting on action denial capability.} Heatmap of accuracy across three setups: the \textit{Standard MCQ}, \textit{Explicit Denial} with a clarifying verification cue, and \textit{No Distractor} with primary competing action removed. Performance improves as contextual ambiguity is reduced. Best model in each setup is highlighted.
    }
    \label{fig:eval_heatmap}
\end{figure}

\noindent \textbf{Effect of Progressive Reduction of Contextual Confusion.} 
To better understand denial failures, we evaluate two modified MCQ setups that progressively reduce contextual ambiguity. In \emph{Explicit Denial}, the generic “None” option is replaced with a statement that the scene may contain cues for the action but the action itself is not occurring, nudging models toward verification rather than context matching. In \emph{Primary Distractor Removed}, the main competing action label is replaced with a random distractor, removing the strongest alternative. As shown in \cref{fig:eval_heatmap}, accuracy increases from the Standard \textit{Overall-AD} MCQ to Explicit Denial and improves further when the primary distractor is removed, often nearing performance on positive actions. This suggests that models can deny actions when ambiguity is reduced, but still struggle in realistic settings where strong contextual cues and plausible distractors remain present. However, in the real-world we cannot simply remove competing actions or simplify the label space, so models must learn to deny actions even when multiple plausible interpretations remain. Details of setups are in supplementary.

\begin{figure}[]
\begin{minipage}[c]{0.59\textwidth}
    \centering
    {
      \includegraphics[width=\linewidth]{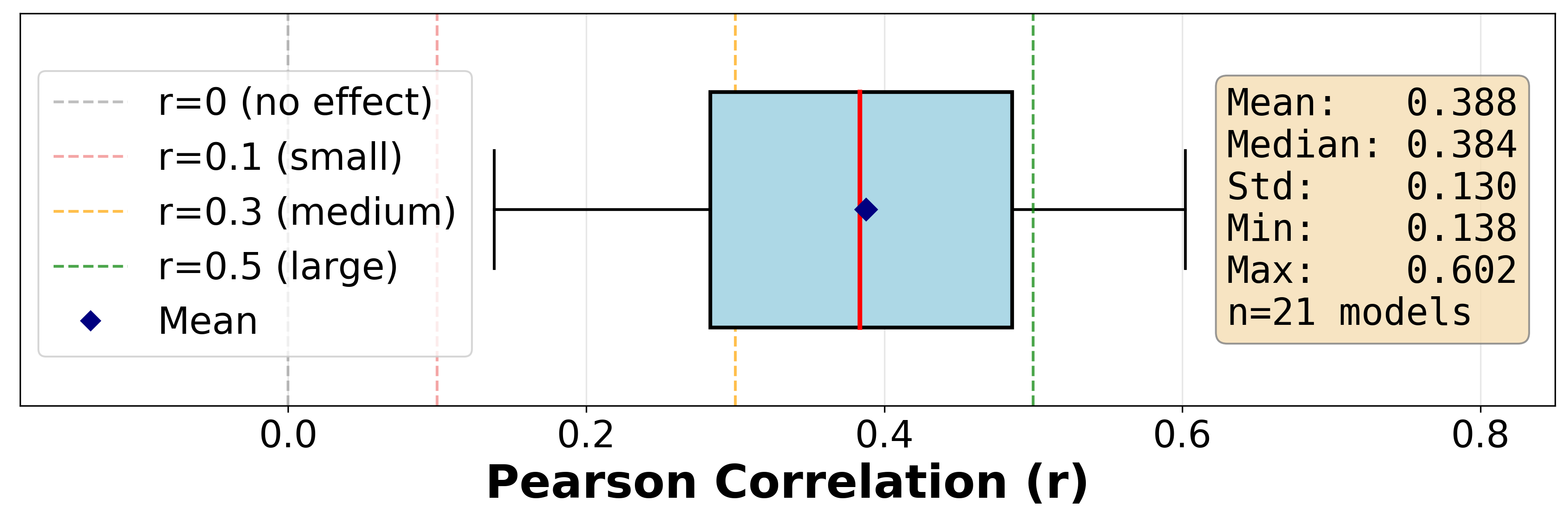}
    }
\end{minipage}\hfill
\begin{minipage}[c]{0.39\textwidth}
    \caption{
        \textbf{Distribution of dual-failure correlations.} Pearson's $r$ coefficient correlates each model's shortcut learning error with its sycophancy error showing a positive correlation ($r > 0$).
    }
    \label{fig:correlation_boxplot}
\end{minipage}

\end{figure}

\noindent\textbf{Correlating shortcut learning and sycophancy.} 
To test whether shortcut learning and sycophancy are independent failure modes or manifestations of a single underlying bias, we construct an additional \textbf{Binary Yes/No} setup that directly measures sycophancy (e.g., ``Is a person or group of people applying makeup to their eyes?''). We then compute a per-video Pearson correlation between shortcut learning and sycophancy, each measured as an error rate: the \textbf{MCQ error rate} is the frequency of failing to select the correct ``None'' option in the Standard MCQ, and the \textbf{sycophancy rate} is the frequency of incorrectly answering ``yes'' in the binary task. As shown in \cref{fig:correlation_boxplot}, all models exhibit a positive correlation between these two errors, with a mean correlation of 0.388 and $p < 0.01$ for every model, indicating at least a medium association by Cohen's conventions \cite{cohenbook} for most models. This relationship is slightly stronger on \textbf{Type 2} clips (same context, different motion; mean $r = 0.409$) than on \textbf{Type 1} clips (context only; mean $r = 0.367$), and most models follow this pattern. Moreover, for every model we observe $P(\text{yes} \mid \text{MCQ wrong}) > P(\text{yes} \mid \text{MCQ correct})$ and $P(\text{yes} \mid \text{primary distractor}) > P(\text{yes} \mid \text{random distractor})$, usually by $0.2$-$0.6$, showing that sycophantic answers concentrate precisely on cases where the model has already latched onto the misleading action hypothesis. This provides evidence that shortcut learning and sycophancy are tightly correlated: models infer actions from contextual cues rather than verifying the defining motion.

\noindent \textbf{General vs. Thinking models.} \
Our analysis of the negative videos in the UCF101-ActionDenial dataset reveals a counterintuitive pattern: reasoning models generally perform worse than their standard counterparts (except Ovis2.5). Instead of improving robustness, their chains-of-thought often over-index on visual context, confidently selecting the most plausible distractor action or even unrelated options. In many cases they also struggle to properly interpret the denial instruction itself, instead hallucinating descriptions of actions that are not present. This suggests that current ``thinking'' models are poorly calibrated for negative constraints: tending to explain what might be happening rather than justify that nothing is.

\section{Improving Action Denial through Causal Graph}
UCF101-AD reveals a fundamental limitation in current MLLMs: they often equate contextual cues with the occurrence of an action. These models operate under a correlational paradigm, implicitly modeling the presence of context with the existence of action and thus lack the capacity for denial. To address this limitation, we move from context-based reasoning to \textbf{causal reasoning}. We treat an action as the outcome of a structured combination of factors including participants, environment, interactions, and most importantly, motion dynamics. We therefore introduce \textbf{CausalAct}, which organizes these elements into a causal structure and encourages the model to verify the required evidence before predicting an action, rather than relying on contextual shortcuts.

\subsection{CausalAct}
\begin{figure}[]
\begin{minipage}[c]{0.44\textwidth}
    \centering
    {
      \includegraphics[width=0.9\linewidth]{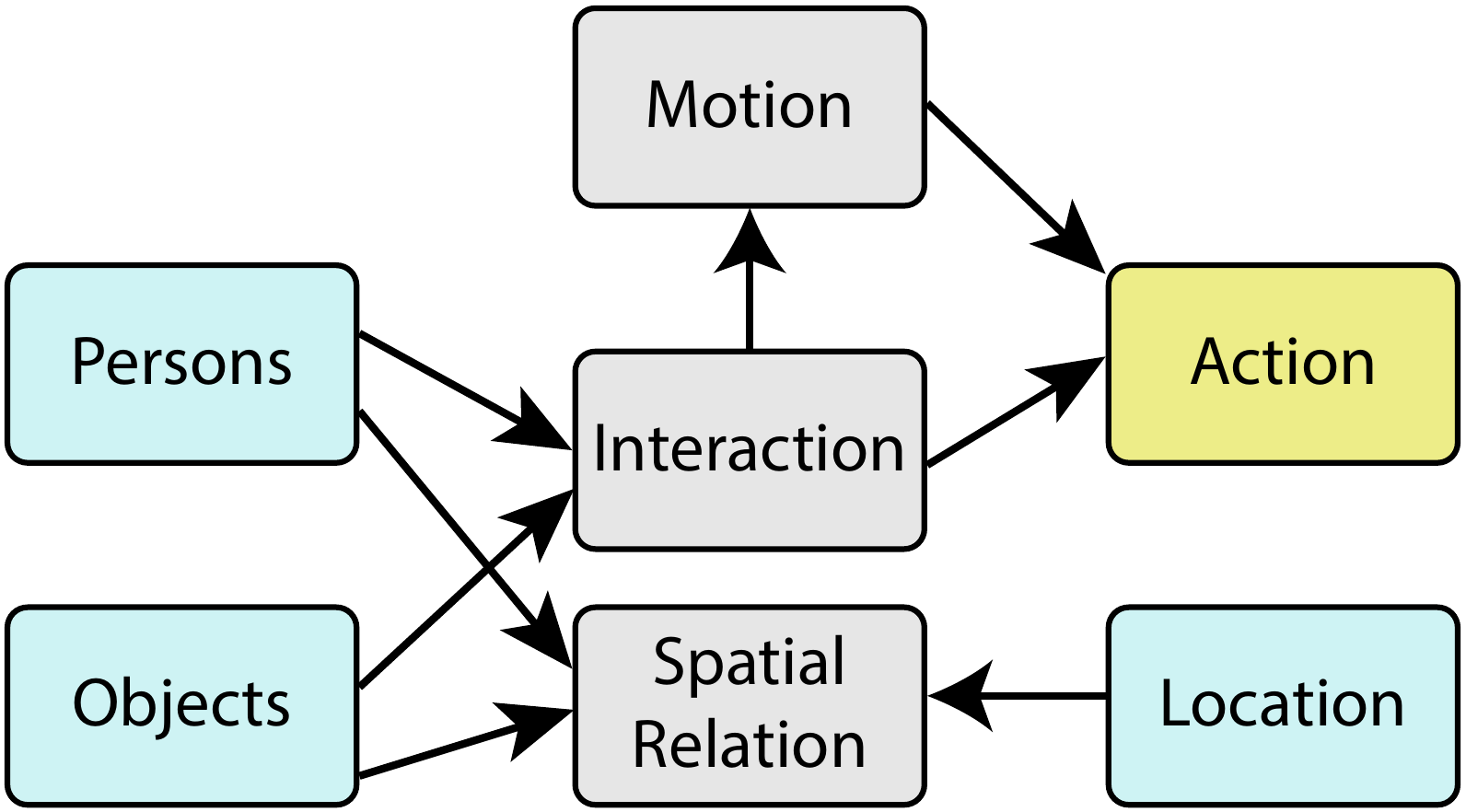}
    }
\end{minipage}\hfill
\begin{minipage}[c]{0.54\textwidth}
    \caption{\textbf{\textit{CausalAct showing the components of an action scene:}} Persons (P), Objects (O), and Location (L) are contextual nodes; Spatial Relation (S) and Interaction (I) are relational nodes; Motion (M) is the dynamic node; and Action (A) is the final activity label.
    }
    \label{fig:scenegenome_graph}
\end{minipage}
\end{figure}

\noindent \textbf{Causal Framework.} We model the evidence required for recognizing an action using a directed acyclic graph (DAG) that organizes scene elements into contextual, relational, and dynamic components. Inspired by structured video representations such as ActionGenome \cite{actiongenome}, \textbf{CausalAct} decomposes an action scene into the following variables: contextual nodes for persons (P), objects (O), and location (L); relational nodes capturing spatial relations (S) and interactions (I) between entities; a dynamic node representing motion (M); and an action node (A) representing the final activity label (\cref{fig:scenegenome_graph}). 
The dependencies between these variables follow a causal hierarchy:
\begin{equation}
    S \leftarrow f_S(P,O,L), \quad I \leftarrow f_I(P,O), \quad M \leftarrow f_M(I), \quad A \leftarrow f_A(I,M),
\end{equation}
where each $f(\cdot)$ denotes an abstract mapping from the video elements that may influence the corresponding variable.

Intuitively, contextual elements (persons, objects, and environment) determine the spatial configuration of the scene, which enables potential interactions. These interactions generate motion patterns, and only when the appropriate motion is observed should the action be inferred. This structure explicitly prevents shortcut reasoning where models predict actions directly from context ($A\leftarrow L$ or $A\leftarrow O$). Instead, the model must verify the presence of interaction and motion before confirming the action.
The graph also accommodates different action types: for person-centric actions, reasoning may proceed directly from motion to action, whereas object-centric actions require both interaction and motion as prerequisites.

\noindent\textbf{Operationalizing CausalAct.} To operationalize the proposed causal structure within MLLMs, we translate CausalAct into a structured reasoning framework using the CausalAct Prompt, a natural language instantiation of the graph. The prompt describes the causal variables and their dependencies, guiding the model to examine contextual elements (persons, objects, and environment), relational evidence (spatial relations and interactions), and motion dynamics before predicting the action. Rather than acting as a strict rule-based module, CausalAct provides visual-grounding guidance that encourages the model to organize the observed evidence through the causal graph rather than shortcutting from scene/object context. By explicitly structuring this reasoning process, it encourages the model to verify whether the required causal chain is present and to deny the action when critical evidence, particularly the defining motion, is absent. However, models with weak vision-language grounding may struggle to reliably follow this structured reasoning.

To address this, we introduce an auxiliary finetuning stage that teaches models the dependencies encoded in \textbf{CausalAct}. For each video, we construct a graph over the variables $(P,O,L,S,I,M,A)$ and automatically generate graph-based question-answer tasks probing node identities, edge relationships, causal paths, and property consistency. 
These questions are derived directly from the graph structure and combined with synthetic distractors, forcing the model to reason over the causal dependencies rather than memorizing action labels. Importantly, the tasks \textbf{avoid exposing the ground-truth action to prevent label leakage}. \textit{Learning} these graph reasoning tasks encourages models to internalize the causal relationships between context, interaction, motion, and action, resulting in improved ability to correctly deny actions in \textit{Action-Denial} scenarios. Details of finetuning are provided in Supplementary.

\subsection{Evaluating CausalAct}
\label{sec:solution_results}

\begin{figure}[]
\centering
{
  \includegraphics[width=\linewidth]{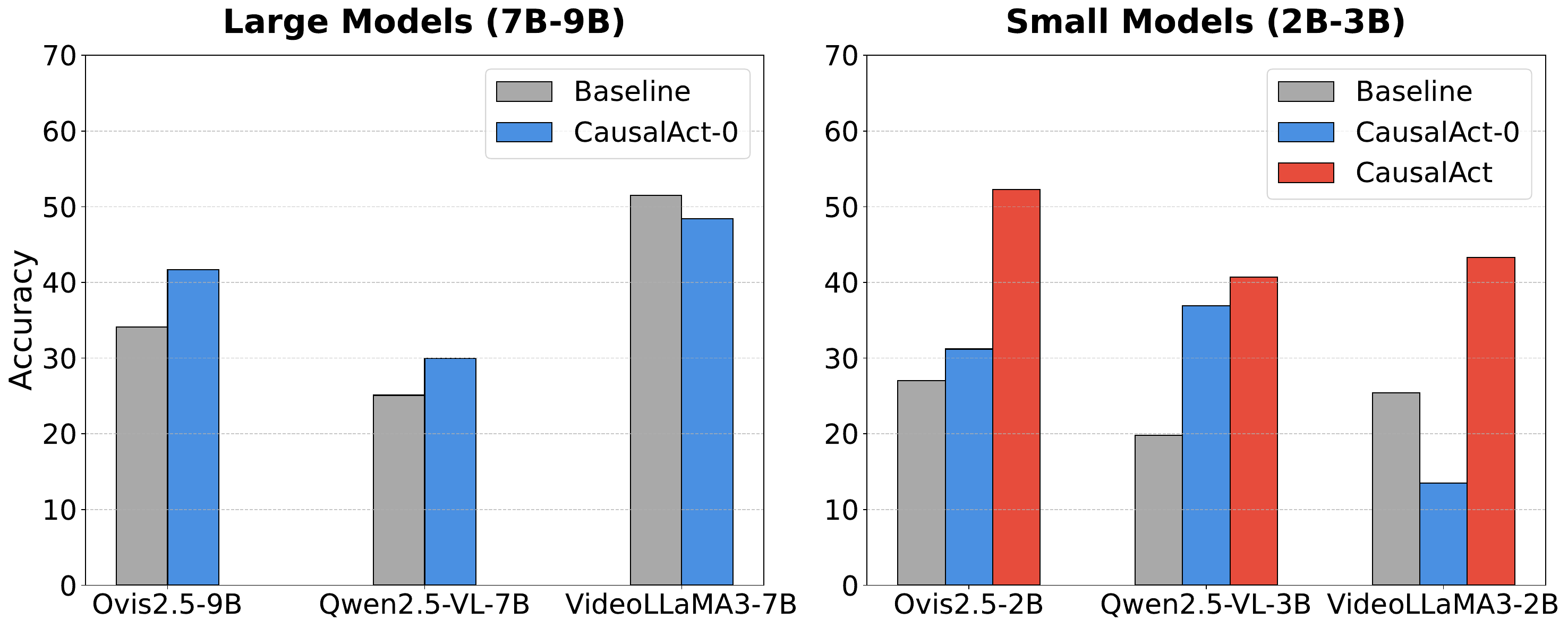}
}
\caption{\textbf{Comparison of baseline, CausalAct-0, and CausalAct.} Left: Larger models with 7B-9B parameters (only zero-shot); right: small models with 2B-3B parameters.
}
\label{fig:model_performance_solution}
\end{figure}

\noindent \textbf{Effect of CausalAct in Action Denial.} We first evaluate the graph in a zero-shot setting (\textit{CausalAct-0}). As shown in \cref{fig:model_performance_solution}, models with stronger language reasoning (Ovis2.5, Qwen2.5-VL) leverage it to better reject false positives, whereas VideoLLaMA3 often fails to follow the causal instructions and can even degrade, indicating that prompting alone is insufficient under weak vision-language alignment. 

\begin{figure}[]
\centering
{
  \includegraphics[width=\linewidth]{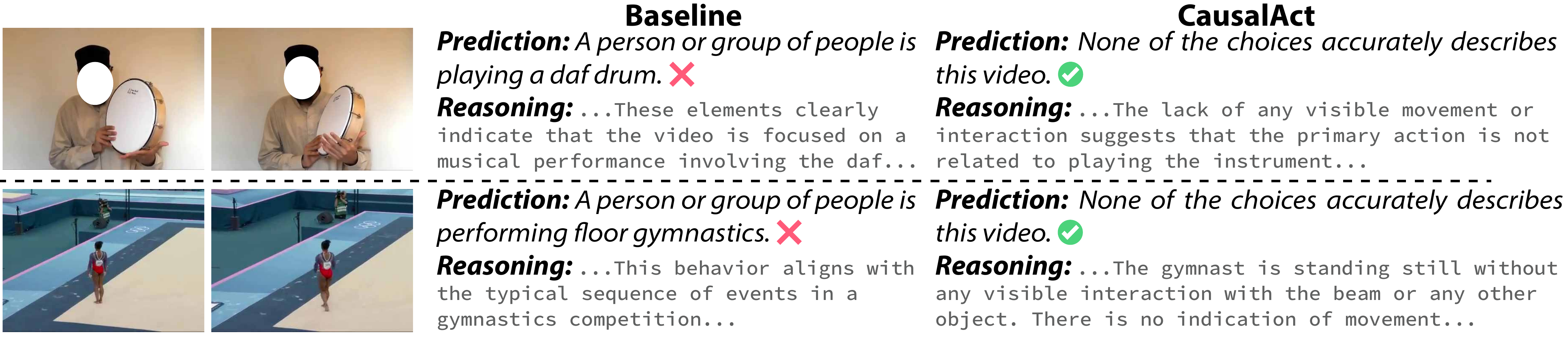}
}
\caption{\textbf{\textit{Qualitative comparison of the baseline vs. CausalAct.}} The base Qwen2.5-VL-3B-Instruct takes shortcuts based on contextual cues in the scene, whereas CausalAct correctly leverages the graph structure to verify the defining motion, enabling it to deny the non-existent action for both \textit{Not Playing Daf} and \textit{Not Floor Gymnastics}. 
}
\label{fig:base_vs_finetuned}
\end{figure}

\noindent We then finetune the smaller models on graph-based reasoning questions derived from CausalAct, which substantially improves denial accuracy on the UCF101-AD \textit{Action-Denial} test set and shifts behavior from contextual shortcutting to denying actions when the defining motion is absent. Although trained only on abstract graph-structure questions, not negative labels, the models acquire a transferable skill of checking the causal chain before asserting an action, as illustrated in \cref{fig:base_vs_finetuned}.

\begin{wraptable}[10]{r}{6.1cm}
\scriptsize
\vspace{-1.1cm}
\centering
\caption{\textbf{\textit{Generalization performance on external action datasets.}}
We compare Ovis2.5-9B's baseline denial accuracy with CausalAct-0.}
\label{tab:generalization_results}
\resizebox{\linewidth}{!}{
\begin{tabular}{lccc}
\toprule
\textbf{Dataset} & \textbf{Base} $\uparrow$ & \textbf{CausalAct-0} $\uparrow$ & \textbf{$\Delta$} \\
\midrule
UCF101 \cite{ucf101}       & 78.7 & 86.1 & +7.4 \\
K400 \cite{kinetics}       & 80.4 & 91.9 & +11.5 \\
HMDB51 \cite{hmdb51}       & 51.2 & 74.3 & +23.1 \\
SSv2 \cite{ssv2}           & 16.0 & 37.4 & +21.4 \\
Diving48 \cite{diving48}   & 48.2 & 70.9 & +22.7 \\
FineGym99 \cite{finegym}   & 71.6 & 87.4 & +15.8 \\
\bottomrule
\end{tabular}}
\end{wraptable}

\noindent\textbf{Generalizability to other datasets:}
To evaluate whether CausalAct-0 transfers beyond our benchmark, we apply it to Ovis2.5-9B on additional action datasets by replacing the ground-truth action with ``None of the choices...''. As shown in \cref{tab:generalization_results}, CausalAct-0 improves denial accuracy across all evaluated datasets, including UCF101 \cite{ucf101}, K400 \cite{kinetics}, HMDB51 \cite{hmdb51}, and SSv2 \cite{ssv2}. We further evaluate on fine-grained motion datasets, Diving48 \cite{diving48} and FineGym99 \cite{finegym}, where verifying an action often requires subtle temporal motion, interaction, or ordering cues. These gains suggest that causal reasoning helps the model verify whether the defining motion or outcome of a candidate action truly occurs, improving action denial across diverse dataset distributions.

\subsection{Ablation Studies}
\begin{figure}[h!]
\centering
{
  \includegraphics[width=\linewidth]{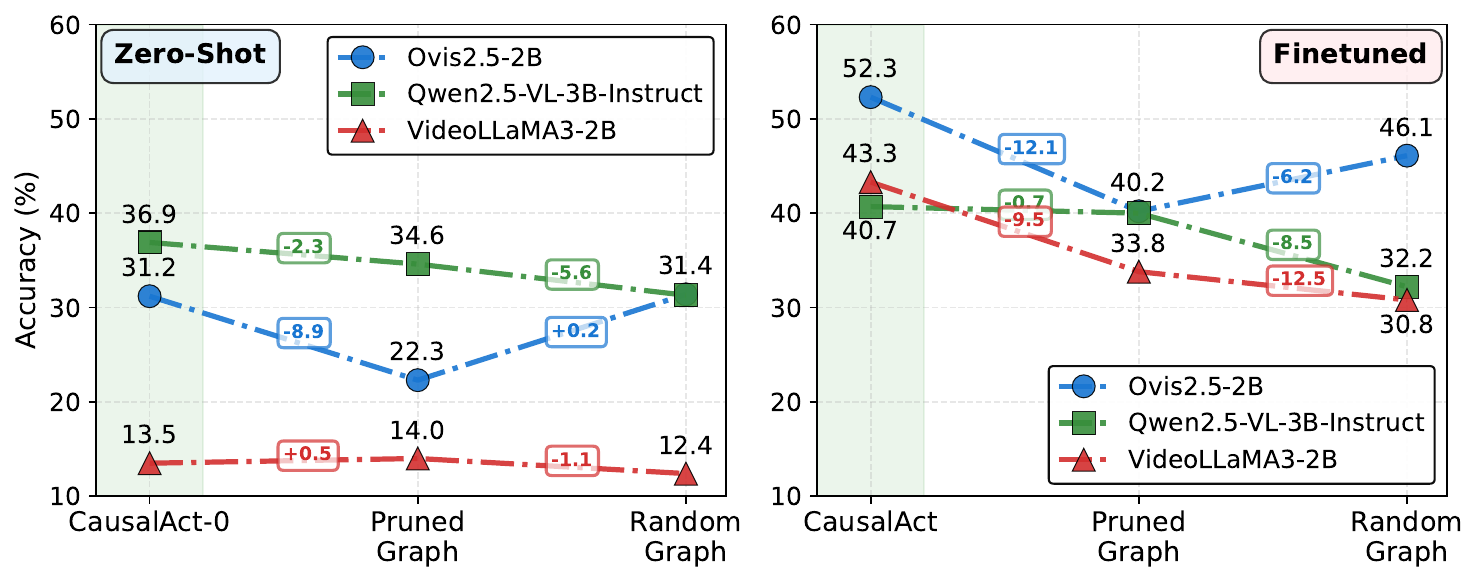}
}
\caption{\textbf{Ablation study on graph structure:} We compare CausalAct to a Pruned Graph (P-O-I-A only) and a Random Graph (shuffled edges). Numeric annotations quantify decline (-) or improvement (+) relative to CausalAct. 
}
\label{fig:model_graph_performance}
\end{figure}

\noindent\textbf{Does graph structure matter?} To test whether performance gains come from the specific causal structure or just any arbitrary graph, we ablate CausalAct by comparing it to a Pruned Graph (only P, O, I, A nodes) and, a Random Graph (all nodes, but shuffled edges). \cref{fig:model_graph_performance} shows that in zero-shot setting (CausalAct-0), models show limited sensitivity to these variants, suggesting that without additional training, they struggle to fully understand the graph’s structure. After finetuning, however, accuracy drops in most cases when the causal dependencies are disrupted, indicating that models have learned to utilize the full structure.  These trends suggest that the prompt only serves as an interface for expressing the causal graph to the MLLM. Finetuning with structured representations helps models internalize and use the intended causal graph for reasoning, while cases with minimal drops hint that they weakly rely on the graph or do not parse it. 

\begin{figure}[]
\begin{minipage}[c]{0.49\textwidth}
    \centering
    {
      \includegraphics[width=\linewidth]{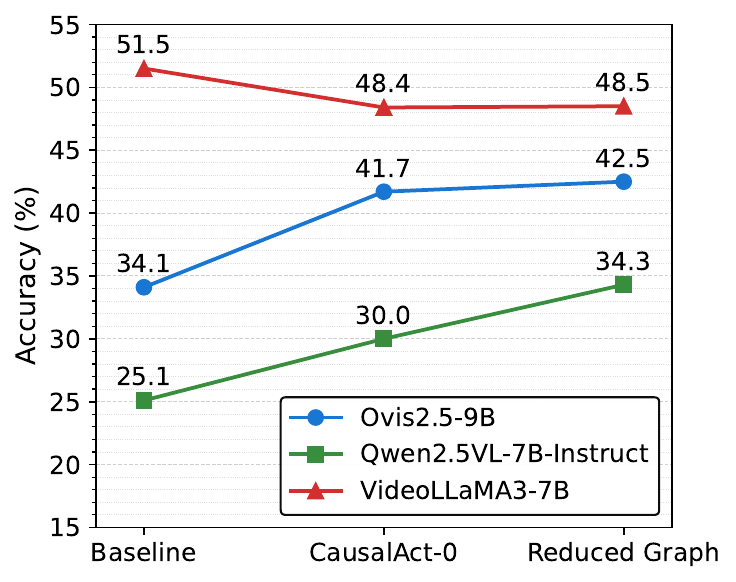}
    }
\end{minipage}\hfill
\begin{minipage}[c]{0.49\textwidth}
    \caption{\textbf{\textit{Effect of removing object nodes for person-only actions.}} Baseline and CausalAct-0 are evaluated on all \textit{Action-Denial} videos, while the Reduced Graph is evaluated on body and location-based actions. Comparable performance shows the causal structure remains effective without explicit object-interaction reasoning.}
    \label{fig:model_performance_no_object_solution}
\end{minipage}
\end{figure}

\noindent\textbf{Is the object node necessary for person-only actions?} To test this, we evaluate a reduced graph that omits object and interaction nodes (mapping only Person, Location, Spatial Relation, Motion, and Action) on a subset of 25 purely body and location-based negatives (e.g., Not Baby Crawling, Not Cliff Diving). As shown in \cref{fig:model_performance_no_object_solution}, zero-shot accuracy with both the full and reduced graphs follows the same trend and matches or exceeds the baseline, indicating that CausalAct’s gains largely stem from organizing causal relations among core scene elements, even when no explicit object reasoning is required.

\noindent\textbf{Is it just language tuning?} We ask whether graph conceptualization finetuning simply teaches new text patterns or requires deeper vision-language alignment. We compare finetuning the \textbf{full model} vs. finetuning \textbf{only the LLM and projector}, freezing the vision encoder. We observed that restricting updates to the visual stack consistently hurts performance, with accuracy dropping by roughly 15\% on average. This systematic degradation indicates that our method is not simply encouraging prompt-following. To effectively ground the causal graph, the vision encoder must also be updated so that visual evidence about the presence or absence of the action is utilized by the MLLM.

\section{Conclusion}
We introduce \textbf{UCF101-AD}, a benchmark targeting a critical failure of modern MLLMs: denying false actions under strong contextual cues. Our experiments on 20 MLLMs show that models excel at positive action recognition but perform poorly on our \textit{Action-Denial} test set (most scoring $<50\%$), revealing a dual bias: a perceptual reliance on static shortcuts and a linguistic tendency toward agreeableness. To guide models toward causal reasoning, we propose \textbf{CausalAct} and accompanying graph-conceptualization finetuning tasks, which improve their capacity for denial. We intend UCF101-AD to serve as a tool for advancing causal reasoning and teaching models to deny.

\section*{Acknowledgements}
We gratefully acknowledge Vibhav Vineet [Microsoft Research, USA], Shahzad Ahmad [Norwegian University of Science and Technology, Norway], and Sukalpa Chanda [Østfold University College, Norway] for their valuable support and contributions to this work.

%
%
\bibliographystyle{splncs04}
\bibliography{main}

\newpage
\renewcommand{\thesection}{\Alph{section}}
\renewcommand{\thesubsection}{\thesection.\arabic{subsection}}
\setcounter{section}{0}

{\centering
{\large \textbf{Supplementary Material of \\
Learning to Deny: Action Denial in Multimodal Large Language Models}}
\par}

This supplementary material provides additional details that complement the main paper.
\begin{itemize}
    \item \cref{sec:detailed_imp_and_results} and \cref{sec:additional_analysis} presents additional implementation details and analysis.
    \item \cref{sec:dataset} provides additional details on dataset curation, evaluation setups, and model prompts.
    \item \cref{sec:causal_graph} elaborates on the CausalAct framework, including the full prompt structure, graph-based finetuning questions.
    \item \cref{sec:neg_classes} lists the semantic descriptions used to construct textual distractors for multiple-choice evaluation and the full set of UCF101-AD \textit{Action-Denial} classes.
    \item \cref{sec:limits} and \cref{sec:ethics} discusses future scope, ethical considerations and dataset use.
\end{itemize}

\section{Additional Implementation Details}
\label{sec:detailed_imp_and_results}
We evaluate 20 diverse MLLMs, spanning both standard architectures and five reasoning-oriented models. Four of these reasoning models are directly derived from standard counterparts: Ovis2.5-9B with reasoning mode enabled; Kimi-VL-A3B-Thinking, the reasoning counterpart to Kimi-VL-A3B-Instruct~\cite{kimi-vl}; and Video-R1-7B~\cite{video-r1} and Lumian-VLR-7B-Thinking~\cite{lumian-vlr}, both built on Qwen2.5-VL. In the zero-shot setting, most models are evaluated on a single NVIDIA RTX A6000 (48GB), while a small number of larger models require a single A100 (80GB).

For fine-tuning with the train set based on CausalAct, we use a single A100 (80GB) GPU for the 2B/3B variants of Ovis2.5, Qwen2.5-VL, and VideoLLaMA3. The training setups share several common aspects: DeepSpeed-based single-GPU training, bfloat16 precision, gradient accumulation over 16 steps, cosine learning-rate scheduling, and gradient checkpointing.

\section{Additional Analysis}
\label{sec:additional_analysis}

\cref{tab:evaluation_setups} shows the detailed scores of the discussed setups.

\begin{table}[]
\scriptsize
\centering
\caption{Accuracy (\%) of UCF101-AD \textit{Overall-AD} and alternative evaluation setups showing the effect of progressive hinting. \textit{Binary} reformulates the task as a yes/no decision. \textbf{Best} and \underline{second-best} performance are highlighted.}
\label{tab:evaluation_setups}
\begin{tabular}{l |c c c| c}
\toprule
\textbf{Model} & \textbf{Overall-AD} \textuparrow & \textbf{Explicit} \textuparrow & \textbf{Primary Distractor} \textuparrow & \textbf{Binary} \textuparrow \\
 &  & \textbf{Denial} & \textbf{Removed} &  \\
\midrule
Ovis2.5-9B \cite{ovis2_5} & 34.1 & {\textbf{74.7}} & 93.9 & 61.1 \\
Ovis2-8B \cite{ovis} & 26.4 & 65.9 & 93.2 & 51.9 \\
InternVideo2.5\_Chat-8B \cite{internvideo2_5} & 18.4 & 44.8 & 91.9 & 45.7 \\
InternVL2.5-8B \cite{internvl2_5} & 31.6 & 54.2 & 97.3 & 44.9 \\
Qwen2.5-VL-7B-Instruct \cite{qwen-vl2_5} & 25.1 & 57.7 & 94.9 & 61.3 \\
VideoLLaMA3-7B \cite{videollama3} & {\textbf{51.5}} & 51.9 & \underline{98.4} & 36.8 \\
VideoChat-Flash-Qwen2-7B \cite{videochat} & 29.4 & 44.4 & 91.4 & 40.7 \\
Oryx-7B \cite{oryx} & 22.1 & 58.3 & 83.2 & {\textbf{79.5}} \\
Valley-Eagle-7B \cite{valley-eagle} & 11.1 & 47.2 & 85.3 & 60.5 \\
LLaVA-Video-7B-Qwen2 \cite{llavavideo} & 29.9 & \underline{71.4} & 91.0 & 71.9 \\
Kimi-VL-A3B-Instruct \cite{kimi-vl} & 29.6 & 47.0 & 87.4 & 68.1 \\
Ovis2.5-2B \cite{ovis2_5} & 27.0 & 55.2 & 85.1 & 55.3 \\
Qwen2.5-VL-3B-Instruct \cite{qwen-vl2_5} & 19.8 & 41.2 & 83.1 & 66.6 \\
VideoLLaMA3-2B \cite{videollama3} & 25.4 & 34.7 & 89.8 & 57.5 \\
Qwen2.5-VL-72B-Instruct \cite{qwen-vl2_5} & \underline{45.7} & 57.6 & \textbf{98.9} & 67.1 \\
GPT4o-mini \cite{gpt4omini} & 21.5 & 46.7 & 93.8 & 64.9 \\
\midrule
\multicolumn{5}{c}{\textit{Reasoning Models}} \\
\midrule
Ovis2.5-9B (thinking) \cite{ovis2_5} & 40.4 & 61.2 & 97.6 & 60.6 \\
Video-R1-7B \cite{video-r1} & 14.6 & 64.1 & 89.0 & \underline{76.7} \\
Kimi-VL-A3B-Thinking \cite{kimi-vl} & 20.9 & 38.8 & 94.4 & 57.4 \\
ARC-Hunyuan-Video-7B \cite{arc-hunyuan} & 14.7 & 32.9 & 42.8 & 43.7 \\
Lumian-VLR-7B-Thinking \cite{lumian-vlr} & 10.3 & 22.2 & 38.7 & 73.6 \\
\bottomrule
\end{tabular}
\end{table}

\noindent\textbf{Effect on number of choices:} We further analyze a subset of the higher-performing models under a reduced 4-choice setting, while keeping both the primary distractor and the ``None'' option. As shown in \cref{tab:choices_no}, performance fluctuates when moving from 11 choices to 4 choices: several models show modest gains, whereas others decline. Importantly, despite the reduced answer space, nearly all evaluated models still remain below 50\% accuracy on \textit{Overall-AD}. These results suggest that the difficulty is not primarily driven by the number of answer choices. Instead, the central challenge lies in action denial itself: models struggle to reject the primary distractor when strong contextual cues make the target action appear plausible.

\begin{table}[]
\scriptsize
\centering
\caption{\textbf{Effect of reducing the number of answer choices on UCF101-AD \textit{Overall-AD}.} Reported values are accuracy (\%) in the standard 11-choice setting and a reduced 4-choice setting that retains the primary distractor and the ``None'' option.}
\label{tab:choices_no}
\resizebox{0.6\linewidth}{!}{
\begin{tabular}{l | c c}
\toprule
\textbf{Model} & \textbf{11 choices} \textuparrow & \textbf{4 choices} \textuparrow \\
\midrule
Ovis2.5-9B \cite{ovis2_5}                & 34.1 & 36.4\\
Ovis2-8B \cite{ovis}                     & 26.4 & 26.1\\
InternVL2.5-8B \cite{internvl2_5}        & 31.6 & 34.7\\
Qwen2.5-VL-7B-Instruct \cite{qwen-vl2_5} & 25.1 & 30.5\\
VideoLLaMA3-7B \cite{videollama3}        & 51.5 & 43.7\\
VideoChat-Flash-Qwen2-7B \cite{videochat}& 29.4 & 21.6\\
LLaVA-Video-7B-Qwen2 \cite{llavavideo}   & 29.9 & 30.2\\
Qwen2.5-VL-72B-Instruct \cite{qwen-vl2_5}& 45.7 & 48.4\\
\midrule
\multicolumn{3}{c}{\textit{Reasoning Models}} \\
\midrule
Ovis2.5-9B (thinking) \cite{ovis2_5}     & 40.4 & 48.6\\
\bottomrule
\end{tabular}
}
\end{table}

\noindent\textbf{Effect of scale:}
As shown in \cref{tab:scaling}, increasing model scale generally improves performance on the standard \textit{Overall-AD} setting, but the strength of this effect differs across model families. The Ovis2.5 series exhibits a relatively modest gain. Qwen2.5-VL-Instruct shows progressively larger variants yielding increasingly larger gains. VideoLLaMA3 displays the largest jump among the compared families. Taken together, these results suggest that model scale is helpful for action denial, but only when the underlying architecture is able to translate added capacity into better rejection of contextually plausible yet absent actions.

\begin{table}[]
\scriptsize
\centering
\caption{\textbf{Effect of model scale on \textit{Overall-AD}.} $\Delta$ denotes the accuracy gain in percentage points, relative to the smallest model in each architecture family.}
\label{tab:scaling}
\begin{tabular}{l | c c}
\toprule
\textbf{Model} & \textbf{Overall-AD} \textuparrow & \textbf{$\Delta$}  \\
\midrule
Ovis2.5-2B \cite{ovis2_5}                 & 27.0 &  \\
Ovis2.5-9B \cite{ovis2_5}                 & 34.1 & +7.1 \\
\hline
Qwen2.5-VL-3B-Instruct \cite{qwen-vl2_5}  & 19.8 &  \\
Qwen2.5-VL-7B-Instruct \cite{qwen-vl2_5}  & 25.1 & +5.3 \\
Qwen2.5-VL-72B-Instruct \cite{qwen-vl2_5} & 45.7 & +25.9 \\
\hline
VideoLLaMA3-2B \cite{videollama3}         & 25.4 &  \\
VideoLLaMA3-7B \cite{videollama3}         & 51.5 & +26.1 \\
\bottomrule
\end{tabular}
\end{table}

\noindent\textbf{Computational cost of CausalAct:} To quantify the additional computation introduced by CausalAct, we report GPU inference time in minutes on UCF101-AD while keeping the model, dataset, hardware, frame sampling, output length, and decoding settings fixed, changing only the input prompt. As shown in \cref{tab:computation}, CausalAct-0 increases runtime due to its longer structured prompt, but the overhead remains moderate for most models, and GPU memory usage stays comparable to the baseline. Furthermore, Ovis2.5-9B-Thinking requires 639 minutes to achieve 40.4\% Overall-AD, whereas standard Ovis2.5-9B with CausalAct-0 reaches a comparable denial accuracy in only 109 minutes. This suggests that CausalAct-0 provides a practical way to encourage reasoning without introducing substantial computational overhead.

\begin{table}[]
\scriptsize
\centering
\caption{\textbf{Computational overhead of CausalAct-0 on UCF101-AD.} GPU inference time is reported in minutes for Base and CausalAct-0 under identical settings.}
\label{tab:cost}

\label{tab:computation}
\begin{tabular}{l c | c c}
\toprule
\textbf{Model} & \textbf{Size} & \textbf{Base} & \textbf{CausalAct-0} \\
\midrule
\multirow{2}{*}{Qwen2.5-VL-Instruct \cite{qwen-vl2_5}} 
& 3B & 96 & 112 \\
& 7B & 106 & 125 \\
\midrule
\multirow{2}{*}{Ovis2.5 \cite{ovis2_5}} 
& 2B & 91 & 115 \\
& 9B & 70 & 109 \\
\midrule
\multirow{2}{*}{VideoLLaMA3 \cite{videollama3}} 
& 2B & 40 & 48 \\
& 7B & 53 & 75 \\
\bottomrule
\end{tabular}%
\end{table}

\section{UCF101-AD Benchmark: Detailed Prompt and Evaluation Setups}
\label{sec:dataset}

\begin{figure}[]
    \centering
    \includegraphics[width=\linewidth]{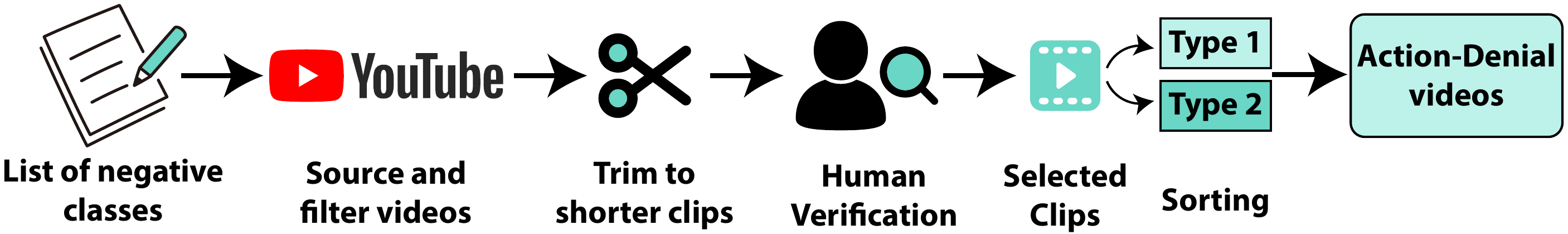}
    \caption{Overview of the video curation pipeline for UCF101-AD \textit{Action-Denial} videos.
    }
    \label{fig:ucf101_neg_data_gen_detail}
\end{figure}

For \textit{Action-Denial} clips, we began by defining negative classes corresponding to the UCF101 actions and querying YouTube with keywords designed to retrieve videos that preserve similar contextual cues, including people, objects, and locations, while excluding the defining motion of the target action. The retrieved videos were then manually trimmed into shorter clips, with an average duration of $\sim$7 seconds. The overall curation procedure is illustrated in \cref{fig:ucf101_neg_data_gen_detail}. To prevent information leakage, clips originating from the same long source video were grouped together and assigned entirely to either the training or test split, following the UCF101 protocol. 

\begin{figure}[]
    \centering
    \includegraphics[width=\linewidth]{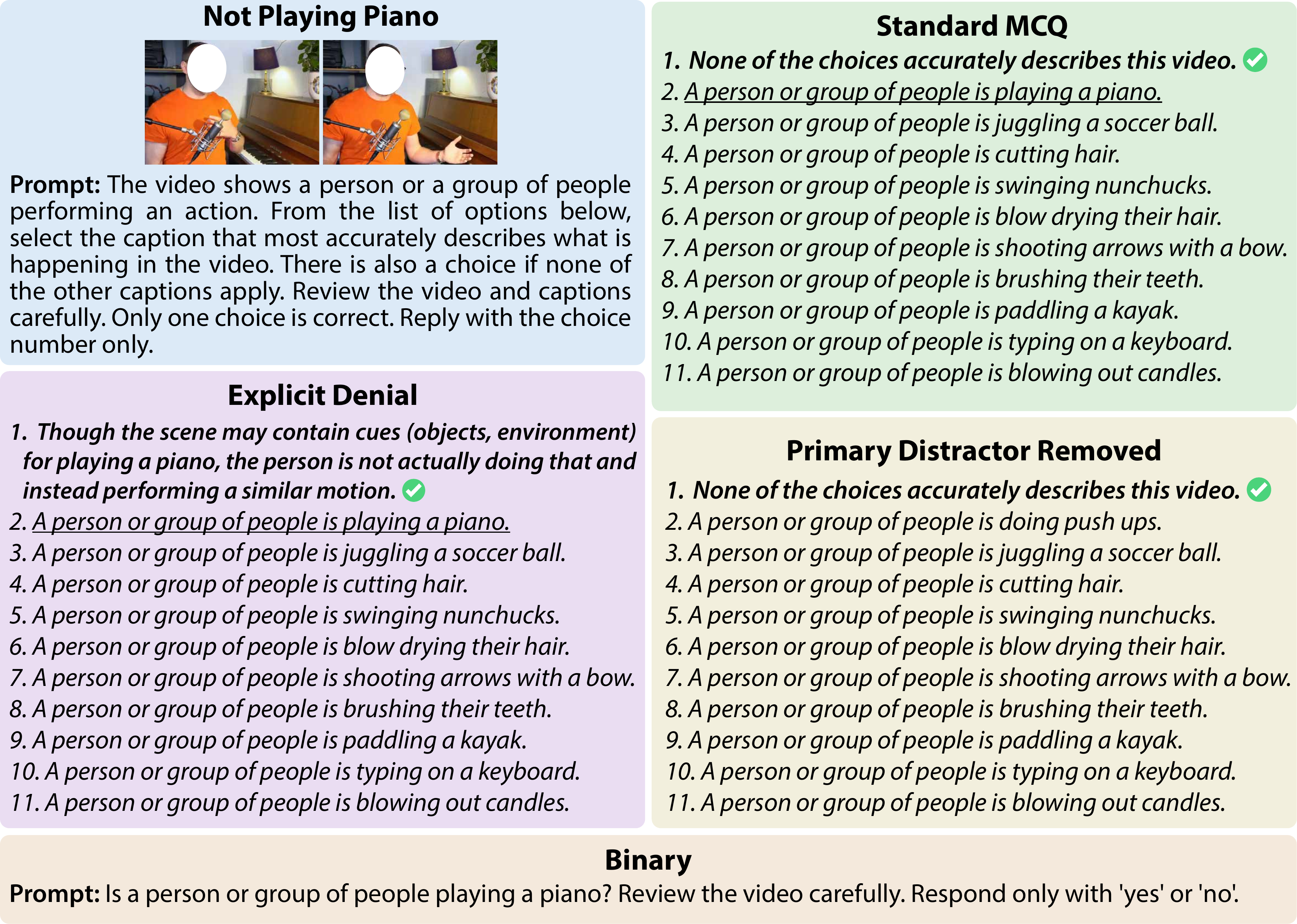}
    \caption{\textbf{Illustration of the evaluation and analytical variants.} \textbf{Standard Overall-AD} includes a \textit{None} option, the primary distractor corresponding to the target action, and additional distractors from other classes. \textbf{Explicit Denial} replaces the \textit{None} option with a natural-language denial statement that explicitly says the target action is not occurring. \textbf{Primary Distractor Removed} replaces the main distractor from the candidate set while keeping the \textit{None} option and the other distractors. \textbf{Binary} reformulates the task as a yes/no question asking whether the target action is present in the video. \textbf{Bold options} highlight the ground truth while the \underline{underlined options} denote the primary distractor.
    }
    \label{fig:ucf101_neg_example_setups}
\end{figure}

Examples of the \textit{Action-Denial} MCQ evaluation and analysis setups, along with the question prompt, are shown in \cref{fig:ucf101_neg_example_setups}. Notably, in the \textit{Action-Denial} setting, all distractor options other than the primary distractor are randomly sampled from unrelated action classes. By contrast, for UCF101~\cite{ucf101} and our \textit{Action-Presence} videos, distractors are selected to be semantically similar to the target action; for example, \textit{playing guitar} may be paired with \textit{playing sitar}. Despite this stronger distractor similarity in the positive setting, models still perform well on those videos. This indicates that the primary difficulty lies not in action recognition itself, but in correctly rejecting an action when contextual evidence strongly implies its presence.

\FloatBarrier

\section{The CausalAct Framework}
\label{sec:causal_graph}

\begin{figure}[]
\centering
\includegraphics[width=\linewidth]{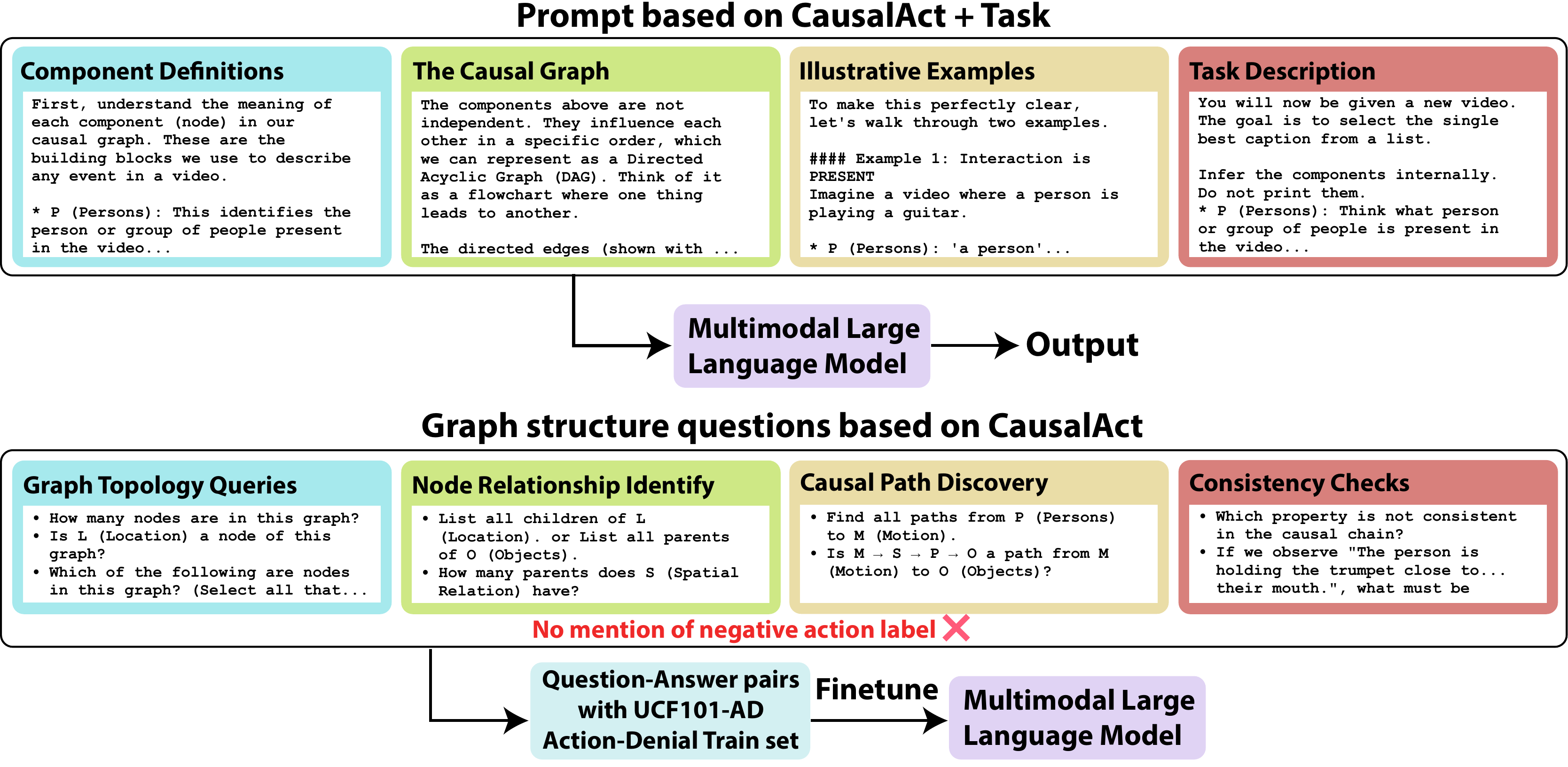}
\caption{\textbf{Overview of CausalAct.} \textbf{Top:} The CausalAct representation organizes video understanding components as nodes in a directed acyclic graph (DAG), which is then used to construct the detailed prompt provided to the MLLM. \textbf{Bottom:} During graph-based question-answer finetuning, \textit{Action-Denial} training video information is converted into templated questions covering graph topology, node relations, causal paths, and consistency constraints. The target action label is never directly exposed; instead, the model is encouraged to internalize the underlying causal reasoning structure.
}
\label{fig:causalact_overview}
\end{figure}

\subsection{Causal Prompting}

We translate CausalAct into a structured natural language prompt. This prompt serves as a comprehensive reasoning instruction for the model. It begins by defining each component (node) of the graph. It articulates the causal dependencies, explaining how each component can be influenced by others and how their interplay ultimately shapes the perceived target action. To reinforce this logic, the prompt includes examples demonstrating both positive and negative scenarios, effectively teaching the model the rules of the causal system. This graph-based prompting aligns with recent efforts to structure inputs for Large Language Models \cite{graphprompts, sgvlm, hyperglm}. As illustrated in \cref{fig:causalact_overview}, this detailed causal prompt is prepended to the user's task query, guiding the model to evaluate the video through this causal lens. The complete prompt structure is detailed below.

\subsection{Detailed CausalAct Prompt}
{\small\ttfamily\raggedright
Your primary task is to act as a causal reasoning engine for video understanding. You will be given a set of observed components from a video. Your goal is to use the provided causal graph structure to logically determine the final action and select the most accurate descriptive caption from a list.

\noindent\textbf{Component Definitions}

First, understand the meaning of each component (node) in our causal graph. These are the building blocks we use to describe any event in a video.

\begin{itemize}
    \item \textbf{P (Persons):} This identifies the person or group of people present in the video. For example, `a man', `a woman', `two children'.
    \item \textbf{O (Objects):} This lists the main inanimate objects that are present in the video. For example, `a guitar', `a basketball', `a chair'.
    \item \textbf{L (Location):} This describes the environment or setting where the event takes place. For example, `a park', `a kitchen', `an office'.
    \item \textbf{S (Spatial Relation):} This describes the positioning of persons and objects relative to each other and the location. It's a snapshot of \textit{where} things are. For example, `a person is standing next to a table'.
    \item \textbf{I (Interaction):} This describes the sustained physical contact or manipulation between a person and an object over a duration. The presence of an interaction is a crucial clue. For example, `a person's hands are on a keyboard' or `a child's feet are on the bicycle pedals'. If there is no contact, the interaction is `None'.
    \item \textbf{M (Motion):} This describes the specific movements and body motions of the persons. It is the dynamic part of the video. For example, `fingers pressing keys', `legs pedaling', `arms swinging'.
    \item \textbf{A (Action):} This is the high-level activity that you must infer. It is the final conclusion based on all the other components. For example, `typing', `biking', `tennis swing'. There might be no relevant action happening at all.
\end{itemize}

\noindent\textbf{The Causal Graph: How Components are Related}

The components above are not independent. They influence each other in a specific order, which we can represent as a Directed Acyclic Graph (DAG). Think of it as a flowchart where one thing leads to another.

The directed edges (shown with right arrows $\rightarrow$ or \textbackslash u2192 character) show the direction of influence. For example, $P \rightarrow I$ means that the presence of a Person (P) is a prerequisite for an Interaction (I) to occur.

The directed edges of the graph are:

$P \rightarrow I$

$O \rightarrow I$

$P \rightarrow S$

$O \rightarrow S$

$L \rightarrow S$

$I \rightarrow M$

$M \rightarrow A$

$I \rightarrow A$

Here are the causal relationships (the directed edges) explained in detail:

\begin{itemize}
    \item $P \rightarrow I$ and $O \rightarrow I$ (How Interaction is formed): For an Interaction (I) to happen, there must be both a Person (P) to perform the interaction and an Object (O) to be interacted with. The interaction is the direct result of the person engaging with the object.
    \item $P \rightarrow S$, $O \rightarrow S$, and $L \rightarrow S$ (How Spatial Relation is formed): The Spatial Relation (S) is determined by where the Persons (P) and Objects (O) are located within the Location (L).
    \item $I \rightarrow M$ (Interaction causes Motion): The specific way a person Interacts (I) with an object over time is what causes the Motion (M) we see. For example, the interaction of `hands on a guitar' may cause the motion of `fingers strumming strings'.
    \item $I \rightarrow A$ and $M \rightarrow A$ (How the final Action is determined): The final Action (A) is a direct result of both the Interaction (I) and the Motion (M). The Interaction tells you \textit{what} is being engaged, and the Motion tells you \textit{how} it is being engaged. Together, they define the action.
\end{itemize}

\noindent\textbf{Illustrative Examples}

To make this perfectly clear, let's walk through two examples.

\textbf{Example 1: Interaction is PRESENT}

Imagine a video where a person is playing a guitar.

\begin{itemize}
    \item P (Persons): `a person'
    \item O (Objects): `a guitar'
    \item L (Location): `a room'
    \item Causal Analysis:
    \begin{enumerate}
        \item The `person' (P) and `guitar' (O) exist in the `room' (L).
        \item Their Spatial Relation (S) is: `The person is sitting and holding the guitar'.
        \item Because the person is holding the guitar, a direct Interaction (I) occurs: `The person's hands are on the guitar's neck and strings'.
        \item This interaction may lead to a specific Motion (M): `Fingers are pressing on the frets and the other hand is strumming the strings'.
        \item Therefore, the combination of the Interaction (I) (`hands on guitar') and the Motion (M) (`strumming') leads to the undeniable conclusion for Action (A): `playing guitar'.
    \end{enumerate}
\end{itemize}

\textbf{Example 2: Interaction is ABSENT}

Now, imagine a video of the same person and guitar, but the person is not playing it.

\begin{itemize}
    \item P (Persons): `a person'
    \item O (Objects): `a guitar'
    \item L (Location): `a room'
    \item Causal Analysis:
    \begin{enumerate}
        \item The `person' (P) and `guitar' (O) still exist in the `room' (L).
        \item However, the Spatial Relation (S) is different: `The person is sitting on a chair, and the guitar is leaning against the wall next to them'.
        \item Because the person is not touching the guitar, the Interaction (I) is: `None'.
        \item Since there is no interaction, no relevant Motion (M) is generated. The observed motion might be `The person is sitting still'.
        \item Without a direct Interaction (I) or a related Motion (M) involving the guitar, the Action (A) cannot be `playing guitar'.
    \end{enumerate}
\end{itemize}

\noindent\textbf{Your Task}

You will now be given a new video. The goal is to select the single best caption from a list.

Infer the components internally. Do not print them.

\begin{itemize}
    \item \textbf{P (Persons):} Think what person or group of people is present in the video.
    \item \textbf{O (Objects):} Think what main inanimate objects are present in the video.
    \item \textbf{L (Location):} Think about the environment or setting where the event takes place in.
    \item \textbf{S (Spatial Relation):} Think about the positioning of persons and objects relative to each other and the location.
    \item \textbf{I (Interaction):} Carefully observe and think about any sustained interaction (physical contact or manipulation) between a person and an object over a duration. There might be no interaction at all.
    \item \textbf{M (Motion):} Carefully observe and think about the specific movements and body motions of the persons.
    \item \textbf{A (Action):} ??? Unknown, you must infer this.
\end{itemize}

Use the causal relationships (directed edges) and the component definitions from above to decide the Action. Reason with these components in your head. Do not output your reasoning.

From the list of options below, select the caption that most accurately describes the action happening in the video. There is also a choice if none of the other captions apply. Review the video and captions carefully. Only one choice is correct. Reply with the choice number only.

\textbf{Output format:} reply with the choice number only.
}

\subsection{Adapting models for CausalAct}

We propose an \textbf{auxiliary finetuning stage}. The objective of this stage is \textit{not} to teach action recognition, but to instil the model with an \textbf{axiomatic understanding} of the CausalAct graph's properties. We construct a set of question-answer pairs grounded in the CausalAct structure based on our training videos. This dataset teaches the model the principles of causal reasoning specific to our graph, analogous to methods that teach causal axioms \cite{axiomatictraining}.

The dataset comprises 14 question types across four main categories:
\paragraph{I. Graph Structure Queries:}
This category assesses the model's foundational knowledge of the Directed Acyclic Graph (DAG) topology, including component existence and valid connectivity.
\begin{enumerate}
    \item \textbf{Node Count Query:} Asks for the total number of nodes in the graph. Since the graph structure is fixed with seven components ($P, O, L, S, I, M, A$), this serves as a static verification that the model has internalized the complete set of causal variables.
    
    \item \textbf{Single Node Membership Check (Binary):} Poses a yes/no question regarding the existence of a specific node (e.g., ``Is V (Viewpoint and Camera) a node of this graph?''). It tests the distinction between valid causal nodes and distractor attributes like lighting or audio.
    
    \item \textbf{Multiple Node Selection:} Presents a mixed list of valid nodes and distractors, requiring the model to select all legitimate components. This evaluates the ability to filter the full set of graph nodes from irrelevant noise in a multiple-choice format.
    
    \item \textbf{Edge Count Query:} Asks for the total number of directed edges in the graph. This verifies knowledge of the fixed causal skeleton, which contains exactly eight directed connections.
    
    \item \textbf{Single Edge Validity Check (Binary):} A verification task where the model must accept valid directed edges (e.g., $P \to I$) and reject non-existent or reversed edges. This tests specific knowledge of pairwise causal links.
    
    \item \textbf{Multiple Edge Selection:} A multiple-choice task requiring the identification of all valid edges from a mixed list of real and synthetic connections. This probes edge-level discrimination and directionality.
\end{enumerate}

\paragraph{II. Node Relationship Questions:}
These tasks examine local graph connectivity, focusing on immediate parent-child relationships and the cardinality of direct causes and effects.
\begin{enumerate}
    \setcounter{enumi}{6} 
    \item \textbf{Parent/Child Listing:} Selects a specific node and asks for an exhaustive list of its immediate parents or children. This tests the ability to retrieve the full adjacency list for a given component (e.g., identifying that $I$ is caused by both $P$ and $O$).
    
    \item \textbf{Relationship Cardinality Check:} Asks for the count of parents or children for a specific node (e.g., ``How many parents does M have?''). This requires the model to summarize incoming or outgoing connections without necessarily listing them.
    
    \item \textbf{Relationship Validity Check (Binary):} Presents a specific pair of nodes and a relationship type (parent or child) for verification. This probes fine-grained understanding of immediate adjacency (e.g., confirming $A$ is a child of $M$).
\end{enumerate}

\paragraph{III. Path Discovery Tasks:}
This category evaluates global graph reasoning, requiring the model to trace multi-hop connections across the DAG.
\begin{enumerate}
    \setcounter{enumi}{9} 
    \item \textbf{Full Path Enumeration:} Selects start and end nodes and asks for every valid path connecting them (e.g., finding paths from $P$ to $A$). The model must traverse the graph recursively to identify all valid causal routes, such as $P \to I \to M \to A$ and $P \to I \to A$.
    
    \item \textbf{Specific Path Validity Check:} Presents a concrete sequence of nodes and asks if it constitutes a valid path. Negative samples are generated by stitching together unconnected nodes or creating cycles, testing the model's ability to validate trajectory consistency.
\end{enumerate}

\paragraph{IV. Property Consistency Checks:}
These questions integrate semantic video content with the graph structure, testing causal logic applied to specific textual descriptions.
\begin{enumerate}
    \setcounter{enumi}{11} 
    \item \textbf{Distractor Property Identification:} One component's description is replaced with a property from a conflicting action class (e.g., swapping the ``interaction'' from a typing video with one from a biking video). The model must identify which node is semantically inconsistent with the rest of the causal chain.
    
    \item \textbf{Causal Antecedent Identification:} Samples a causal link and asks what earlier property is required given a later observation (e.g., ``If we observe [Motion Description], what must be present earlier?''). This tests the understanding that downstream effects logically imply specific upstream causes.
    
    \item \textbf{Reverse Causality Verification:} Asks if an earlier property can be influenced by a later property (e.g., ``Can [Interaction] be influenced by [Action]?''). As the graph is acyclic, the answer is invariably negative; this task enforces the strict temporal and causal directionality of the model's reasoning.
\end{enumerate}

\noindent Each question is randomly sampled to ensure balanced coverage of all reasoning patterns. By finetuning smaller models on this graph reasoning set, we hypothesize they will be better equipped to process and utilize the graph-structured prompts for the downstream task of action denial.
 
As shown in \cref{tab:typewise_performance}, CausalAct improves performance on both Type~1 and Type~2 \textit{Action-Denial} videos across. The gains are consistently larger on Type~2 negatives, where the context remains similar but a different motion is performed, suggesting that graph-based knowledge is particularly effective at helping models distinguish the target action from closely related non-target motions. Improvements on Type~1 also indicate better resistance to context-only cues when the defining motion is absent. Overall, these per-type improvements show that CausalAct strengthens action-denial reasoning.

\begin{table}[]
\scriptsize
\caption{\textbf{Type-wise accuracy (\%) on UCF101-AD \textit{Action-Denial} test videos for baseline model vs. CausalAct.} $\Delta$ denotes absolute gains in percentage points.} 
\label{tab:typewise_performance}
\centering
\begin{tabular}{llccc}
\toprule
\textbf{Model} & \textbf{} & \textbf{Base} & \textbf{CausalAct} & $\Delta$ \\
\midrule
\multirow{3}{*}{Ovis2.5-2B \cite{ovis2_5}}
& Type 1 & 23.1 & 43.5 & +20.4 \\
& Type 2 & 30.4 & 60.1 & +29.7 \\
& Overall & 27.0 & 52.3 & +25.3 \\
\midrule
\multirow{3}{*}{Qwen2.5-VL-3B-Instruct \cite{qwen-vl2_5}}
& Type 1 & 17.2 & 32.8 & +15.6 \\
& Type 2 & 22.1 & 47.6 & +25.5 \\
& Overall & 19.8 & 40.7 & +20.9 \\
\midrule
\multirow{3}{*}{VideoLLaMA3-2B \cite{videollama3}}
& Type 1 & 20.2 & 32.6 & +12.4 \\
& Type 2 & 30.0 & 52.7 & +22.7 \\
& Overall & 25.4 & 43.3 & +17.9 \\
\bottomrule
\end{tabular}
\end{table}

\FloatBarrier

\section{UCF101-AD Classes and Semantic Descriptions}
\label{sec:neg_classes}
To mitigate ambiguity inherent in short class names, each original UCF101 \cite{ucf101} category is rewritten as a clear, descriptive sentence to serve as a robust distractor in the MCQ evaluation.
\begin{enumerate}
    \item \textbf{ApplyEyeMakeup:} A person or group of people is applying makeup to their eyes.
    \item \textbf{ApplyLipstick:} A person or group of people is applying lipstick to their lips.
    \item \textbf{Archery:} A person or group of people is shooting arrows with a bow.
    \item \textbf{BabyCrawling:} A person or group of people is crawling on the floor.
    \item \textbf{BalanceBeam:} A person or group of people is flipping on a balance beam.
    \item \textbf{BandMarching:} A person or group of people is marching in a band.
    \item \textbf{BaseballPitch:} A person or group of people is pitching a baseball.
    \item \textbf{Basketball:} A person or group of people is playing basketball shooting at the basket.
    \item \textbf{BasketballDunk:} A person or group of people is successfully dunking a basketball through the basket.
    \item \textbf{BenchPress:} A person or group of people is performing a bench press.
    \item \textbf{Biking:} A person or group of people is riding a bicycle causing it to move forward.
    \item \textbf{Billiards:} A person or group of people is playing billiards hitting shot with stick.
    \item \textbf{BlowDryHair:} A person or group of people is blow drying their hair.
    \item \textbf{BlowingCandles:} A person or group of people is blowing out candles.
    \item \textbf{BodyWeightSquats:} A person or group of people is doing body weight squats.
    \item \textbf{Bowling:} A person or group of people is bowling a ball down an alley knocking down the pins.
    \item \textbf{BoxingPunchingBag:} A person or group of people is punching a boxing bag vigorously with force.
    \item \textbf{BoxingSpeedBag:} A person or group of people is hitting a speed bag.
    \item \textbf{BreastStroke:} A person or group of people is swimming breaststroke.
    \item \textbf{BrushingTeeth:} A person or group of people is brushing their teeth.
    \item \textbf{CleanAndJerk:} A person or group of people is lifting a barbell with a clean and jerk.
    \item \textbf{CliffDiving:} A person or group of people is diving off a cliff into water.
    \item \textbf{CricketBowling:} A person or group of people is bowling a cricket ball releasing it from the hand.
    \item \textbf{CricketShot:} A person or group of people is batting in cricket hitting the ball.
    \item \textbf{CuttingInKitchen:} A person or group of people is cutting ingredients in a kitchen.
    \item \textbf{Diving:} A person or group of people is diving into water.
    \item \textbf{Drumming:} A person or group of people is playing drums.
    \item \textbf{Fencing:} A person or group of people is engaged in the sport of fencing, using specialized swords, and following formal fencing rules and techniques.
    \item \textbf{FieldHockeyPenalty:} A person or group of people is taking a field hockey penalty shot and scoring successfully.
    \item \textbf{FloorGymnastics:} A person or group of people is performing floor gymnastics.
    \item \textbf{FrisbeeCatch:} A person or group of people is catching a frisbee.
    \item \textbf{FrontCrawl:} A person or group of people is swimming front crawl.
    \item \textbf{GolfSwing:} A person or group of people is swinging a golf club.
    \item \textbf{Haircut:} A person or group of people is cutting hair.
    \item \textbf{Hammering:} A person or group of people is hammering a nail.
    \item \textbf{HammerThrow:} A person or group of people is throwing a hammer in athletics.
    \item \textbf{HandstandPushups:} A person or group of people is doing handstand push ups.
    \item \textbf{HandstandWalking:} A person or group of people is walking on hands in a handstand.
    \item \textbf{HeadMassage:} A person or group of people is giving or receiving a head massage.
    \item \textbf{HighJump:} A person or group of people is performing a high jump over a bar.
    \item \textbf{HorseRace:} A person or group of people is riding in a horse race.
    \item \textbf{HorseRiding:} A person or group of people is riding a horse.
    \item \textbf{HulaHoop:} A person or group of people is spinning a hula hoop.
    \item \textbf{IceDancing:} A person or group of people is performing ice dancing.
    \item \textbf{JavelinThrow:} A person or group of people is throwing a javelin.
    \item \textbf{JugglingBalls:} A person or group of people is juggling balls.
    \item \textbf{JumpingJack:} A person or group of people is doing jumping jacks.
    \item \textbf{JumpRope:} A person or group of people is skipping with a jump rope.
    \item \textbf{Kayaking:} A person or group of people is paddling a kayak.
    \item \textbf{Knitting:} A person or group of people is knitting with needles.
    \item \textbf{LongJump:} A person or group of people is performing a long jump.
    \item \textbf{Lunges:} A person or group of people is doing lunges.
    \item \textbf{MilitaryParade:} A person or group of people is marching in a military parade.
    \item \textbf{Mixing:} A person or group of people is mixing ingredients.
    \item \textbf{MoppingFloor:} A person or group of people is mopping the floor.
    \item \textbf{Nunchucks:} A person or group of people is swinging nunchucks.
    \item \textbf{ParallelBars:} A person or group of people is performing on parallel bars.
    \item \textbf{PizzaTossing:} A person or group of people is tossing pizza dough.
    \item \textbf{PlayingCello:} A person or group of people is playing a cello.
    \item \textbf{PlayingDaf:} A person or group of people is playing a daf drum.
    \item \textbf{PlayingDhol:} A person or group of people is playing a dhol drum.
    \item \textbf{PlayingFlute:} A person or group of people is playing a flute.
    \item \textbf{PlayingGuitar:} A person or group of people is playing a guitar.
    \item \textbf{PlayingPiano:} A person or group of people is playing a piano.
    \item \textbf{PlayingSitar:} A person or group of people is playing a sitar.
    \item \textbf{PlayingTabla:} A person or group of people is playing a tabla.
    \item \textbf{PlayingViolin:} A person or group of people is playing a violin.
    \item \textbf{PoleVault:} A person or group of people is pole vaulting.
    \item \textbf{PommelHorse:} A person or group of people is performing on a pommel horse.
    \item \textbf{PullUps:} A person or group of people is doing pull ups.
    \item \textbf{Punch:} A person or group of people is throwing punches in combat.
    \item \textbf{PushUps:} A person or group of people is doing push ups.
    \item \textbf{Rafting:} A person or group of people is rafting on white water.
    \item \textbf{RockClimbingIndoor:} A person or group of people is climbing indoor rock walls.
    \item \textbf{RopeClimbing:} A person or group of people is climbing a rope.
    \item \textbf{Rowing:} A person or group of people is rowing a boat.
    \item \textbf{SalsaSpin:} A person or group of people is spinning while dancing salsa.
    \item \textbf{ShavingBeard:} A person or group of people is shaving a beard.
    \item \textbf{Shotput:} A person or group of people is putting a shot.
    \item \textbf{SkateBoarding:} A person or group of people is riding a skateboard and moving forward.
    \item \textbf{Skiing:} A person or group of people is skiing and progressing downhill.
    \item \textbf{Skijet:} A person or group of people is riding a jet ski and moving forward.
    \item \textbf{SkyDiving:} A person or group of people is skydiving, free-falling straight downwards risking hitting the ground if there was no parachute.
    \item \textbf{SoccerJuggling:} A person or group of people is juggling a soccer ball.
    \item \textbf{SoccerPenalty:} A person or group of people is taking a soccer penalty kick.
    \item \textbf{StillRings:} A person or group of people is performing on still rings.
    \item \textbf{SumoWrestling:} A person or group of people is wrestling in sumo style.
    \item \textbf{Surfing:} A person or group of people is surfing on waves.
    \item \textbf{Swing:} A person or group of people is swinging on a playground swing.
    \item \textbf{TableTennisShot:} A person or group of people is playing a table tennis shot.
    \item \textbf{TaiChi:} A person or group of people is practicing tai chi performing martial forms with striking and defensive gestures.
    \item \textbf{TennisSwing:} A person or group of people is swinging a tennis racket and hitting the ball.
    \item \textbf{ThrowDiscus:} A person or group of people is throwing a discus.
    \item \textbf{TrampolineJumping:} A person or group of people is jumping on a trampoline.
    \item \textbf{Typing:} A person or group of people is typing on a keyboard.
    \item \textbf{UnevenBars:} A person or group of people is performing on uneven bars.
    \item \textbf{VolleyballSpiking:} A person or group of people is spiking a volleyball causing it to hit the ground.
    \item \textbf{WalkingWithDog:} A person or group of people is walking with a dog and moving forward.
    \item \textbf{WallPushups:} A person or group of people is doing push ups against a wall.
    \item \textbf{WritingOnBoard:} A person or group of people is writing on a board.
    \item \textbf{YoYo:} A person or group of people is playing with a yoyo.
\end{enumerate}

\begin{table*}[]
\scriptsize
\centering
\caption{Full list of UCF101-AD \textit{Action-Denial} classes derived from UCF101 actions.}
\label{tab:neg_classes_list}
\resizebox{\textwidth}{!}{%
\begin{tabular}{|l|l|l|}
\hline
1. Not Apply Eye Makeup & 35. Not Hammering & 69. Not Pommel Horse \\
\hline
2. Not Apply Lipstick & 36. Not Hammer Throw & 70. Not Pull Ups \\
\hline
3. Not Archery & 37. Not Handstand Pushups & 71. Not Punch \\
\hline
4. Not Baby Crawling & 38. Not Handstand Walking & 72. Not Push Ups \\
\hline
5. Not Balance Beam & 39. Not Head Massage & 73. Not Rafting \\
\hline
6. Not Band Marching & 40. Not High Jump & 74. Not Rock Climbing Indoor \\
\hline
7. Not Baseball Pitch & 41. Not Horse Race & 75. Not Rope Climbing \\
\hline
8. Not Basketball & 42. Not Horse Riding & 76. Not Rowing \\
\hline
9. Not Basketball Dunk & 43. Not Hula Hoop & 77. Not Salsa Spin \\
\hline
10. Not Bench Press & 44. Not Ice Dancing & 78. Not Shaving Beard \\
\hline
11. Not Biking & 45. Not Javelin Throw & 79. Not Shotput \\
\hline
12. Not Billiards & 46. Not Juggling Balls & 80. Not Skate Boarding \\
\hline
13. Not Blow Dry Hair & 47. Not Jumping Jack & 81. Not Skiing \\
\hline
14. Not Blowing Candles & 48. Not Jump Rope & 82. Not Skijet \\
\hline
15. Not Body Weight Squats & 49. Not Kayaking & 83. Not Sky Diving \\
\hline
16. Not Bowling & 50. Not Knitting & 84. Not Soccer Juggling \\
\hline
17. Not Boxing Punching Bag & 51. Not Long Jump & 85. Not Soccer Penalty \\
\hline
18. Not Boxing Speed Bag & 52. Not Lunges & 86. Not Still Rings \\
\hline
19. Not Breast Stroke & 53. Not Military Parade & 87. Not Sumo Wrestling \\
\hline
20. Not Brushing Teeth & 54. Not Mixing & 88. Not Surfing \\
\hline
21. Not Clean And Jerk & 55. Not Mopping Floor & 89. Not Swing \\
\hline
22. Not Cliff Diving & 56. Not Nunchucks & 90. Not Table Tennis Shot \\
\hline
23. Not Cricket Bowling & 57. Not Parallel Bars & 91. Not Tai Chi \\
\hline
24. Not Cricket Shot & 58. Not Pizza Tossing & 92. Not Tennis Swing \\
\hline
25. Not Cutting In Kitchen & 59. Not Playing Cello & 93. Not Throw Discus \\
\hline
26. Not Diving & 60. Not Playing Daf & 94. Not Trampoline Jumping \\
\hline
27. Not Drumming & 61. Not Playing Dhol & 95. Not Typing \\
\hline
28. Not Fencing & 62. Not Playing Flute & 96. Not Uneven Bars \\
\hline
29. Not Field Hockey Penalty & 63. Not Playing Guitar & 97. Not Volleyball Spiking \\
\hline
30. Not Floor Gymnastics & 64. Not Playing Piano & 98. Not Walking With Dog \\
\hline
31. Not Frisbee Catch & 65. Not Playing Sitar & 99. Not Wall Pushups \\
\hline
32. Not Front Crawl & 66. Not Playing Tabla & 100. Not Writing On Board \\
\hline
33. Not Golf Swing & 67. Not Playing Violin & 101. Not YoYo \\
\hline
34. Not Haircut & 68. Not Pole Vault & \\
\hline
\end{tabular}}
\end{table*}

\FloatBarrier

\section{Future Scope}
\label{sec:limits}
While CausalAct improves action denial in the current benchmark setting, there remains room to extend this problem to more challenging scenarios. Future work can study action denial in longer and more complex videos, where relevant evidence may be sparse, delayed, or distributed across time, and move beyond closed-set evaluation toward open-world settings with unseen actions and greater semantic diversity. A key challenge in such settings is scalability: complex real-world videos may contain many objects, making it impractical to include all visible entities in the causal graph or prompt. Future variants could therefore incorporate actor-centric filtering, focusing only on objects that are spatially close to, manipulated by, or otherwise interacting with the actor before constructing the denial prompt. It would also be interesting to examine how action denial generalizes to embodied or interactive environments. More broadly, future work can scale the benchmark to more diverse domains and investigate how causal reasoning can be adapted to support robust video understanding in these broader settings.

\section{Ethics and Dataset Use}
\label{sec:ethics}
The dataset is constructed from publicly available YouTube videos released under Creative Commons licenses. The released dataset contains trimmed clips rather than full original videos, and is intended solely for research on action denial/recognition. The dataset is not designed for identity recognition, face analysis, or sensitive attribute inference, and no identity-related labels are separately collected. Given the use of publicly available material, we consider the ethical risk of this dataset to be limited.

\end{document}